\documentclass{article}
\usepackage{ijcai24}

\usepackage{times}
\usepackage{soul}

\usepackage{amsmath,amsfonts,bm}

\def\eqref#1{equation~\ref{#1}}

\def\1{\bm{1}}

\DeclareMathAlphabet{\mathsfit}{\encodingdefault}{\sfdefault}{m}{sl}
\SetMathAlphabet{\mathsfit}{bold}{\encodingdefault}{\sfdefault}{bx}{n}

\usepackage[hidelinks]{hyperref}
\usepackage[utf8]{inputenc}
\usepackage[small]{caption}
\usepackage{url}
\usepackage{graphicx}
\usepackage{amsmath}
\usepackage{amsthm}
\usepackage{tikz}
\usetikzlibrary{calc}
\usetikzlibrary{shapes.geometric}
\graphicspath{{images/jupyterImages/},{images/illustrationVideos/}}
\usepackage{caption}
\usepackage{subcaption}
\usepackage{booktabs}
\usepackage{algorithm}
\usepackage{algorithmic}
\usepackage[switch]{lineno}

\colorlet{lightgray}{gray!70}

\tikzset{
 pics/myTrap/.style 2 args={code={
   \def\width{#1}
   \def\height{#2}
   \draw[rounded corners] ($(0,0)-(\width/2,\height/2)$) -- ++(0,\height) -- ++(\width,-\height/4) -- ++(0,-\height/2) -- cycle;
 }},
 conv/.pic={
 \node[draw,inner sep=0, outer sep=0, minimum width=0.05cm, minimum height=1cm] at (0.0,0) (in){};
  \node[draw,inner sep=0, outer sep=0, minimum width=0.05cm, minimum height=0.7cm] at ($(in.east)+(0.1,0)$) (h){};
  \draw (in.north east) -- (h.north west);
  \draw (in.south east) -- (h.south west);
  \foreach \i in {0,...,2} {
    \node[circle, draw,fill, inner sep=0, outer sep=0,minimum width=0.05cm] at ($(h.west) + (0.2,\i*0.2-0.2)$) (neur\i) {};
    \draw (h.north east) -- (neur\i);
    \draw (h.south east) -- (neur\i);
    \draw (h.east) -- (neur\i);
  }
 },
 timeWarperImg/.pic={
   \def\wid{0.5}
   \node[inner sep=0,outer sep=0,minimum width=\wid cm, minimum height=\wid cm] at (0,0) (box) {};
   \draw (box.south west) -- ++(\wid * 0.33,0.1) -- ++(\wid*0.33,0.3) -- ($(box.north east)-(0.04,0.04)$);
   \draw (box.north west) -- (box.south west) -- (box.south east);
 },
 factorizedDecoderImg/.pic={
  \foreach \i in {0,...,2} {
    \node[circle, draw,fill, inner sep=0, outer sep=0,minimum width=0.05cm] at ($(0.0,0.4 + \i*0.2-0.2)$) (neur\i) {};
  }
  \foreach \i in {0,...,4} {
    \node[circle, draw,fill, inner sep=0, outer sep=0,minimum width=0.05cm] at ($(neur1) + (0.2,\i*0.1-0.2)$) (hneur\i) {};
    \draw (neur2.east) -- (hneur\i);
    \draw (neur1.east) -- (hneur\i);
    \draw (neur0.east) -- (hneur\i);
  }
  \node[draw,inner sep=0, outer sep=0, minimum width=0.5cm, minimum height=0.5cm] at ($(hneur2.east) + (0.5,0)$) (out){\tiny$\mathbf T$};
  \foreach \i in {0,...,4} {
    \draw (out.north west) -- (hneur\i);
    \draw (out.west) -- (hneur\i);
    \draw (out.south west) -- (hneur\i);
  }

  \foreach \i in {1} {
    \node[circle, draw,fill, inner sep=0, outer sep=0,minimum width=0.05cm] at ($(0.2,-0.4 + \i*0.2-0.2)$) (tneur\i) {};
  }
  \foreach \i in {0,...,4} {
    \node[circle, draw,fill, inner sep=0, outer sep=0,minimum width=0.05cm] at ($(tneur1) + (0.2,\i*0.1-0.2)$) (thneur\i) {};
    \draw (tneur1.east) -- (thneur\i);
  }
  \node[draw,inner sep=0, outer sep=0, minimum width=0.1cm, minimum height=0.5cm] at ($(thneur2.east) + (0.3,0)$) (tout){\tiny$g$};
  \foreach \i in {0,...,4} {
    \draw (tout.north west) -- (thneur\i);
    \draw (tout.west) -- (thneur\i);
    \draw (tout.south west) -- (thneur\i);
  }

  \node[inner sep=0, outer sep=0] at ($0.5*(tout) + 0.5*(out) + (0.0,0.0)$) (times){$\boldsymbol\times$};
 }
}

\date{}
\title{TimewarpVAE: Simultaneous Time-Warping and\\
Representation Learning of Trajectories}

\author{Travers Rhodes \And Daniel D. Lee \\
Cornell Tech\\
New York, NY 10044, USA \\
\texttt{\{tsr42,ddl46\}@cornell.edu} \\
}   

\newcommand{\lsep}{0.235}

\begin{document}

\maketitle

\begin{abstract}
  Human demonstrations of trajectories are an important source of training data for many machine learning problems. However, the difficulty of collecting human demonstration data for complex tasks makes learning efficient representations of those trajectories challenging. For many problems, such as for dexterous manipulation, the exact timings of the trajectories should be factored from their spatial path characteristics. In this work, we propose TimewarpVAE, a fully differentiable manifold-learning algorithm that incorporates Dynamic Time Warping (DTW) to simultaneously learn both timing variations and latent factors of spatial variation. We show how the TimewarpVAE algorithm learns appropriate time alignments and meaningful representations of spatial variations in handwriting and fork manipulation datasets. Our results have lower spatial reconstruction test error than baseline approaches and the learned low-dimensional representations can be used to efficiently generate semantically meaningful novel trajectories. We demonstrate the utility of our algorithm to generate novel high-speed trajectories for a robotic arm.
\end{abstract}

\section{Introduction}
\begin{figure}
  \centering
  \begin{tikzpicture}
    \def \vidsep{1.6}
    \def\timevalues{{0.0,0.3,1.1,1.5,1.7}}
    \foreach \picnum in {0,...,4} {%
      \node at ($(\picnum*\vidsep,0)$) (f\picnum) {\includegraphics[width=1.5cm, trim=10 0 10 10, clip]{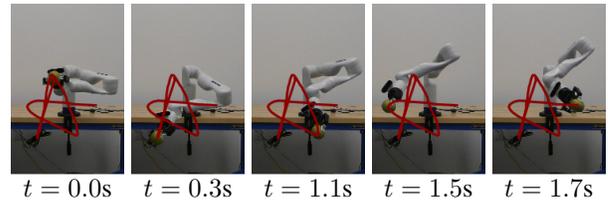}};
      \node at (f\picnum.south) {$t = \pgfmathparse{\timevalues[\picnum]}\pgfmathresult$s};
    }%
\end{tikzpicture}%
\caption{TimewarpVAE learns a low-dimensional latent representation of complex trajectories that explicitly factorizes timing and spatial styles. The Kinova Gen3 robot arm is able to draw various versions of the letter ``A'' more quickly by speeding up or slowing down different parts of the trajectory to obey dynamical mechanical constraints. The resulting end-effector path is overlaid on the images. A video and more details are provided in the Supplemental Materials.}
  \label{fig:trajExecution}
\end{figure}

Continuous trajectories are inherently infinite-dimensional objects that can vary in complex ways in both time and space.  However, in many practical situations, their intrinsic sources of variability can be well-approximated by projection onto a low-dimensional manifold.
For instance, when a human demonstrates trajectories for a robot,
it is useful for the robot to learn to model the most expressive latent factors controlling the functionally-relevant parts of the demonstration trajectories.
For many types of demonstrations, such as in gesture control or quasistatic manipulation,
it is highly advantageous to explicitly separate the exact timing of the trajectory from the spatial latent factors.
This work introduces a method that learns such a representation to generate novel fast trajectories for a robot arm, as shown in Fig~\ref{fig:trajExecution}.

Consider the problem of trying to average two samples from a handwriting dataset generated by humans drawing the letter ``A'' in the air \cite{chen_6dmg_2012}.
If we scale two trajectories linearly in time so that their timestamps go from $0$ to $1$,
and then average the two trajectories at each timestep,
the resulting average does not maintain the style of the ``A''s.
This is because the average is taken between parts of the two trajectories that do not naturally correspond.
An example of this averaging, with lines showing examples of averaged points, is shown in Fig.~\ref{fig:naiveAverage}.
Scaling trajectories to have constant speed also doesn't solve the issue (Fig.~\ref{fig:constantSpeedAverage}).
A common approach like dynamic time warping (DTW) \cite{sakoe_dynamic_1978} can lead to unintuitive results.
When averaging these same two trajectories,
DTW only takes in information about these two trajectories 
and does not use contextual information about other examples of the letter ``A'' to better understand how to align the timings of these trajectories.\footnote{
We use the terms ``time-warping,'' ``time alignment,'' and ``registration'' interchangeably.}
In Fig.~\ref{fig:dtwAverage}, we use the \texttt{dtw} package \cite{Giorgino2009} to align 
the trajectories before averaging them at corresponding timesteps.
We see the resulting trajectory is spatially close to the input trajectories, 
but it again does not maintain the style of the ``A''s.

\newcommand\lossDefinitionScaling{0.92}
\newcommand\figylaba{1.9}
\newcommand\figylabb{-0.9}
\newcommand\figylabc{-2.5}
\newcommand\figxlaba{0.2}
\newcommand\figxlabb{0.1}
\newcommand\coff{-0.2}
\newcommand\trajlaba{Traj. 0}
\newcommand\trajlabb{Traj. 1}
\newcommand\trajlabc{Interp.}
\newcommand\legendysep{0.20}
\newcommand\interplx{0.6}
\newcommand\interply{3.47}
\begin{figure*}%
\centering%
\begin{subfigure}[b]{0.20\textwidth}%
\centering%
\resizebox{0.9\textwidth}{!}{%
\begin{tikzpicture}
  \node at (0,0) (bigpic) {\includegraphics[width=\textwidth, trim=20 70 0 0, clip]{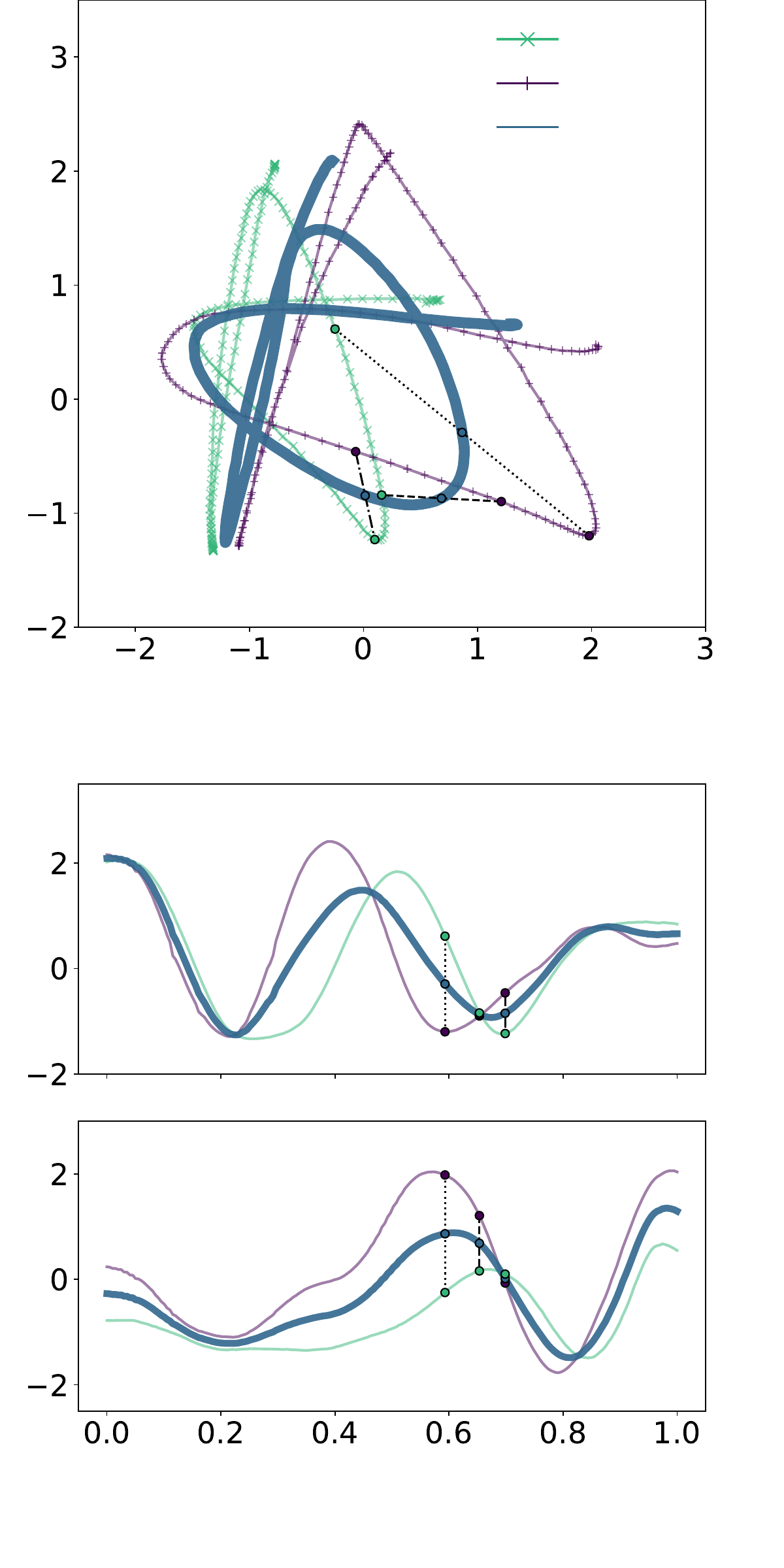}};
    \node[fill=white,minimum width=1cm,minimum height=0.1cm,inner sep=0, outer sep=0] 
      at ($(bigpic.center) + (\coff,\figxlaba)$) {\tiny X Position};
    \node[fill=white,minimum width=1cm,minimum height=0.1cm,anchor=center,rotate=90,inner sep=0, outer sep=0] 
      at ($(bigpic.west) + (0,\figylaba)$) {\tiny Y Position};
    \node[fill=white,minimum width=1cm,minimum height=0.1cm,inner sep=0, outer sep=0] 
      at ($(bigpic.south)+(\coff,\figxlabb)$) {\tiny Canonical Time};
    \node[fill=white,minimum width=1.3cm,minimum height=0.1cm,anchor=center,rotate=90,inner sep=0, outer sep=0] 
      at ($(bigpic.west) + (0,\figylabb)$) {\tiny Y Position};
    \node[fill=white,minimum width=1.3cm,minimum height=0.1cm,anchor=center,rotate=90,inner sep=0, outer sep=0] 
      at ($(bigpic.west) + (0,\figylabc)$) {\tiny X Position};
  \node[anchor=north west,font={\fontsize{5pt}{8}\selectfont}] at (\interplx,\interply) {\trajlaba};
  \node[anchor=north west,font={\fontsize{5pt}{8}\selectfont}] at (\interplx,\interply-\legendysep) {\trajlabb};
  \node[anchor=north west,font={\fontsize{5pt}{8}\selectfont}] at (\interplx,\interply-2*\legendysep) {\trajlabc};
\end{tikzpicture}%
}%
\caption{Uniform time scaling}%
\label{fig:naiveAverage}%
\end{subfigure}%
\hfill%
\begin{subfigure}[b]{0.20\textwidth}%
\centering%
\resizebox{0.9\textwidth}{!}{%
\begin{tikzpicture}
  \node at (0,0) (bigpic) {\includegraphics[width=\textwidth, trim=20 70 0 0, clip]{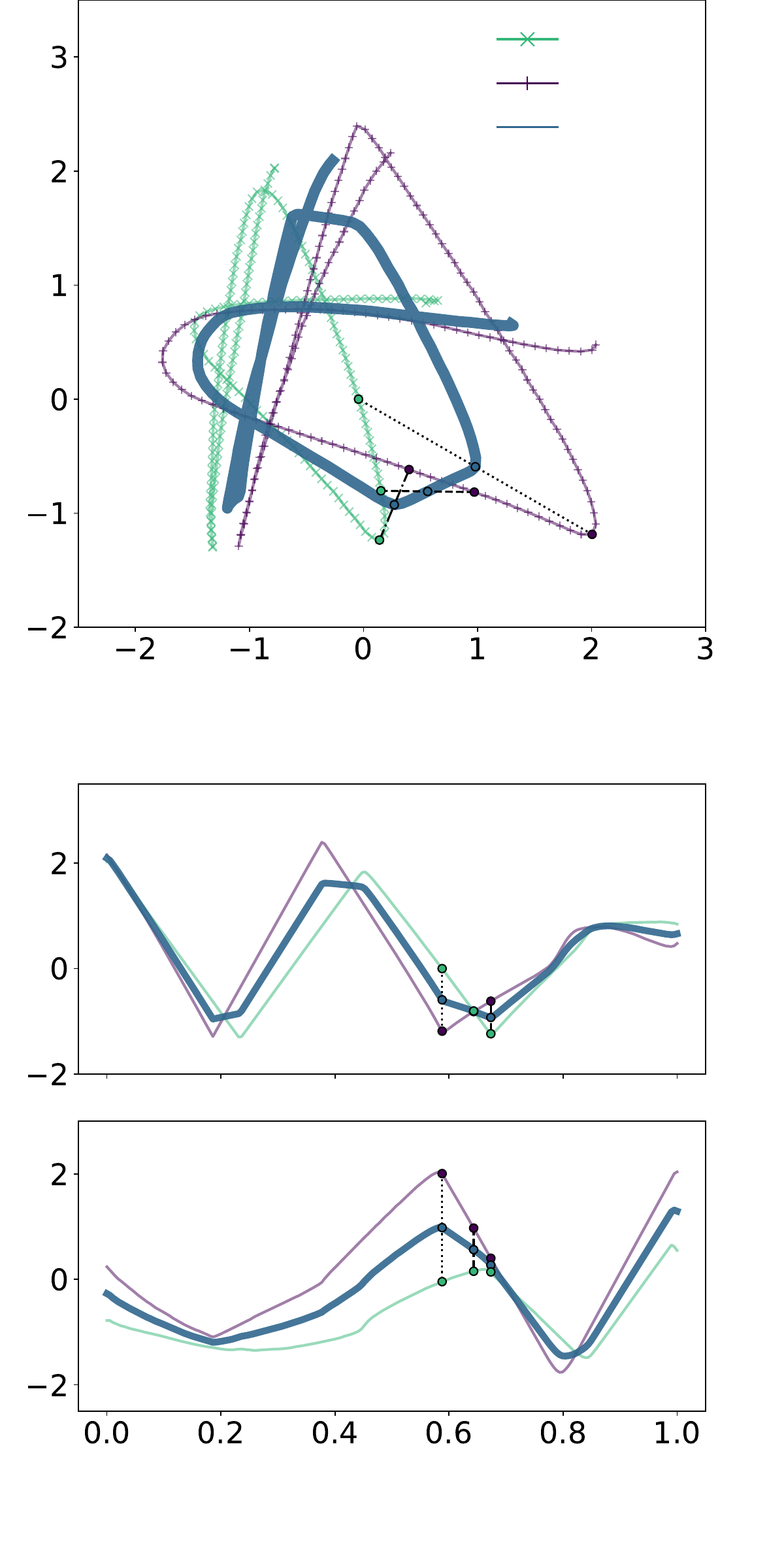}};
    \node[fill=white,minimum width=1cm,minimum height=0.1cm,inner sep=0, outer sep=0] 
      at ($(bigpic.center) + (\coff,\figxlaba)$) {\tiny X Position};
    \node[fill=white,minimum width=1cm,minimum height=0.1cm,anchor=center,rotate=90,inner sep=0, outer sep=0] 
      at ($(bigpic.west) + (0,\figylaba)$) {\tiny Y Position};
    \node[fill=white,minimum width=1cm,minimum height=0.1cm,inner sep=0, outer sep=0] 
      at ($(bigpic.south)+(\coff,\figxlabb)$) {\tiny Canonical Time};
    \node[fill=white,minimum width=1.3cm,minimum height=0.1cm,anchor=center,rotate=90,inner sep=0, outer sep=0] 
      at ($(bigpic.west) + (0,\figylabb)$) {\tiny Y Position};
    \node[fill=white,minimum width=1.3cm,minimum height=0.1cm,anchor=center,rotate=90,inner sep=0, outer sep=0] 
      at ($(bigpic.west) + (0,\figylabc)$) {\tiny X Position};
  \node[anchor=north west,font={\fontsize{5pt}{8}\selectfont}] at (\interplx,\interply) {\trajlaba};
  \node[anchor=north west,font={\fontsize{5pt}{8}\selectfont}] at (\interplx,\interply-\legendysep) {\trajlabb};
  \node[anchor=north west,font={\fontsize{5pt}{8}\selectfont}] at (\interplx,\interply-2*\legendysep) {\trajlabc};
\end{tikzpicture}%
}%
\caption{Constant speed scaling}%
\label{fig:constantSpeedAverage}%
\end{subfigure}%
\hfill%
\begin{subfigure}[b]{0.20\textwidth}%
\centering%
\resizebox{0.9\textwidth}{!}{%
\begin{tikzpicture}
  \node at (0,0) (bigpic) {\includegraphics[width=\textwidth, trim=20 70 0 0, clip]{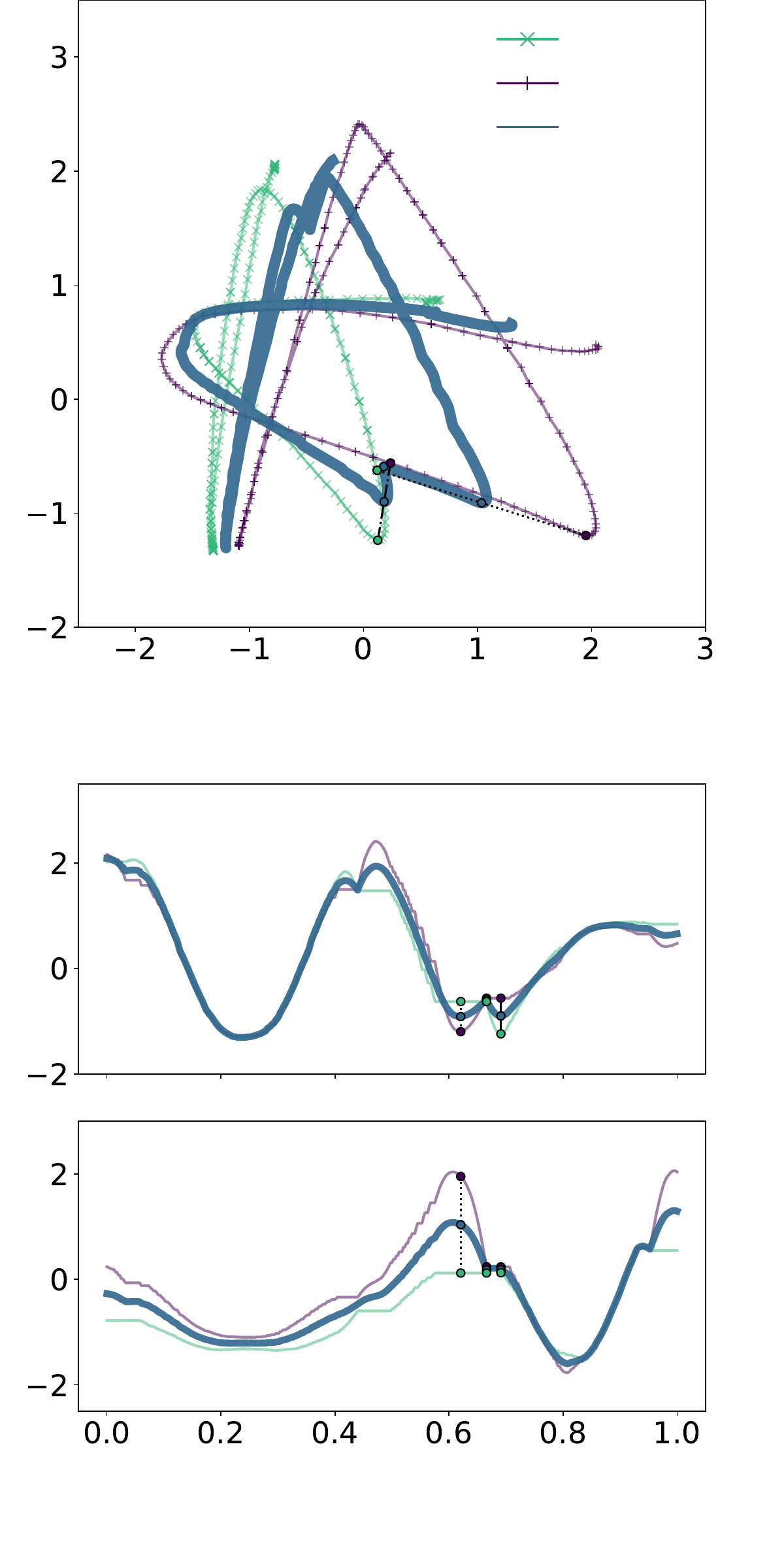}};
    \node[fill=white,minimum width=1cm,minimum height=0.1cm,inner sep=0, outer sep=0] 
      at ($(bigpic.center) + (\coff,\figxlaba)$) {\tiny X Position};
    \node[fill=white,minimum width=1cm,minimum height=0.1cm,anchor=center,rotate=90,inner sep=0, outer sep=0] 
      at ($(bigpic.west) + (0,\figylaba)$) {\tiny Y Position};
    \node[fill=white,minimum width=1cm,minimum height=0.1cm,inner sep=0, outer sep=0] 
      at ($(bigpic.south)+(\coff,\figxlabb)$) {\tiny Canonical Time};
    \node[fill=white,minimum width=1.3cm,minimum height=0.1cm,anchor=center,rotate=90,inner sep=0, outer sep=0] 
      at ($(bigpic.west) + (0,\figylabb)$) {\tiny Y Position};
    \node[fill=white,minimum width=1.3cm,minimum height=0.1cm,anchor=center,rotate=90,inner sep=0, outer sep=0] 
      at ($(bigpic.west) + (0,\figylabc)$) {\tiny X Position};
  \node[anchor=north west,font={\fontsize{5pt}{8}\selectfont}] at (\interplx,\interply) {\trajlaba};
  \node[anchor=north west,font={\fontsize{5pt}{8}\selectfont}] at (\interplx,\interply-\legendysep) {\trajlabb};
  \node[anchor=north west,font={\fontsize{5pt}{8}\selectfont}] at (\interplx,\interply-2*\legendysep) {\trajlabc};
\end{tikzpicture}%
}%
\caption{DTW time alignment}%
\label{fig:dtwAverage}%
\end{subfigure}%
\hfill%
\begin{subfigure}[b]{0.20\textwidth}%
\centering%
\resizebox{0.9\textwidth}{!}{%
\begin{tikzpicture}
  \node at (0,0) (bigpic) {\includegraphics[width=\textwidth, trim=20 70 0 0, clip]{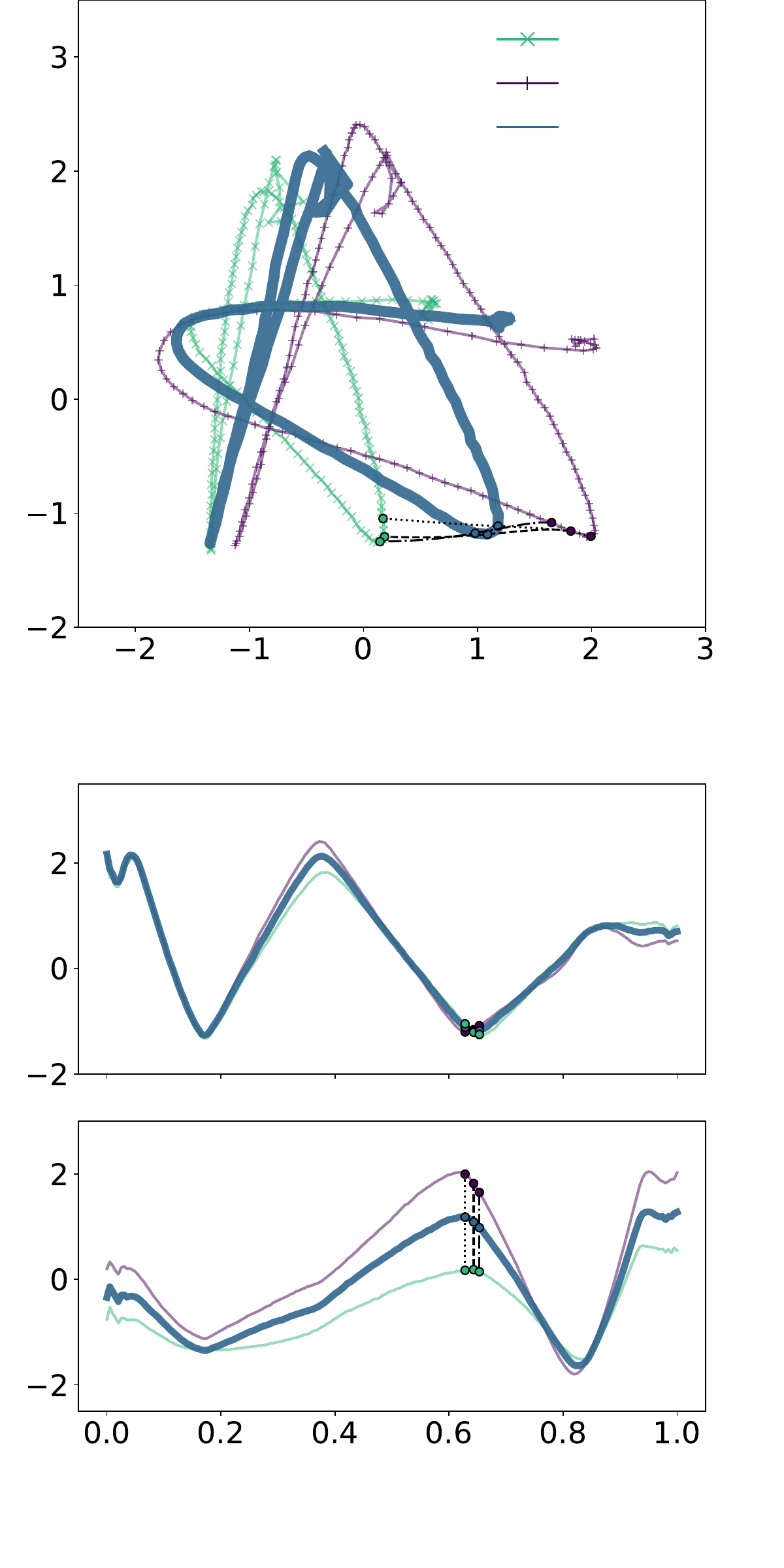}};
    \node[fill=white,minimum width=1cm,minimum height=0.1cm,inner sep=0, outer sep=0] 
      at ($(bigpic.center) + (\coff,\figxlaba)$) {\tiny X Position};
    \node[fill=white,minimum width=1cm,minimum height=0.1cm,anchor=center,rotate=90,inner sep=0, outer sep=0] 
      at ($(bigpic.west) + (0,\figylaba)$) {\tiny Y Position};
    \node[fill=white,minimum width=1cm,minimum height=0.1cm,inner sep=0, outer sep=0] 
      at ($(bigpic.south)+(\coff,\figxlabb)$) {\tiny Canonical Time};
    \node[fill=white,minimum width=1.3cm,minimum height=0.1cm,anchor=center,rotate=90,inner sep=0, outer sep=0] 
      at ($(bigpic.west) + (0,\figylabb)$) {\tiny Y Position};
    \node[fill=white,minimum width=1.3cm,minimum height=0.1cm,anchor=center,rotate=90,inner sep=0, outer sep=0] 
      at ($(bigpic.west) + (0,\figylabc)$) {\tiny X Position};
  \node[anchor=north west,font={\fontsize{5pt}{8}\selectfont}] at (\interplx,\interply) {\trajlaba};
  \node[anchor=north west,font={\fontsize{5pt}{8}\selectfont}] at (\interplx,\interply-\legendysep) {\trajlabb};
  \node[anchor=north west,font={\fontsize{5pt}{8}\selectfont}] at (\interplx,\interply-2*\legendysep) {\trajlabc};
\end{tikzpicture}%
}%
\caption{Rate Invariant AE}%
\label{fig:rateInvariantAverage}%
\end{subfigure}%
\hfill%
\begin{subfigure}[b]{0.20\textwidth}%
\centering%
\resizebox{0.9\textwidth}{!}{%
\begin{tikzpicture}
  \node at (0,0) (bigpic) {\includegraphics[width=\textwidth, trim=20 70 0 0, clip]{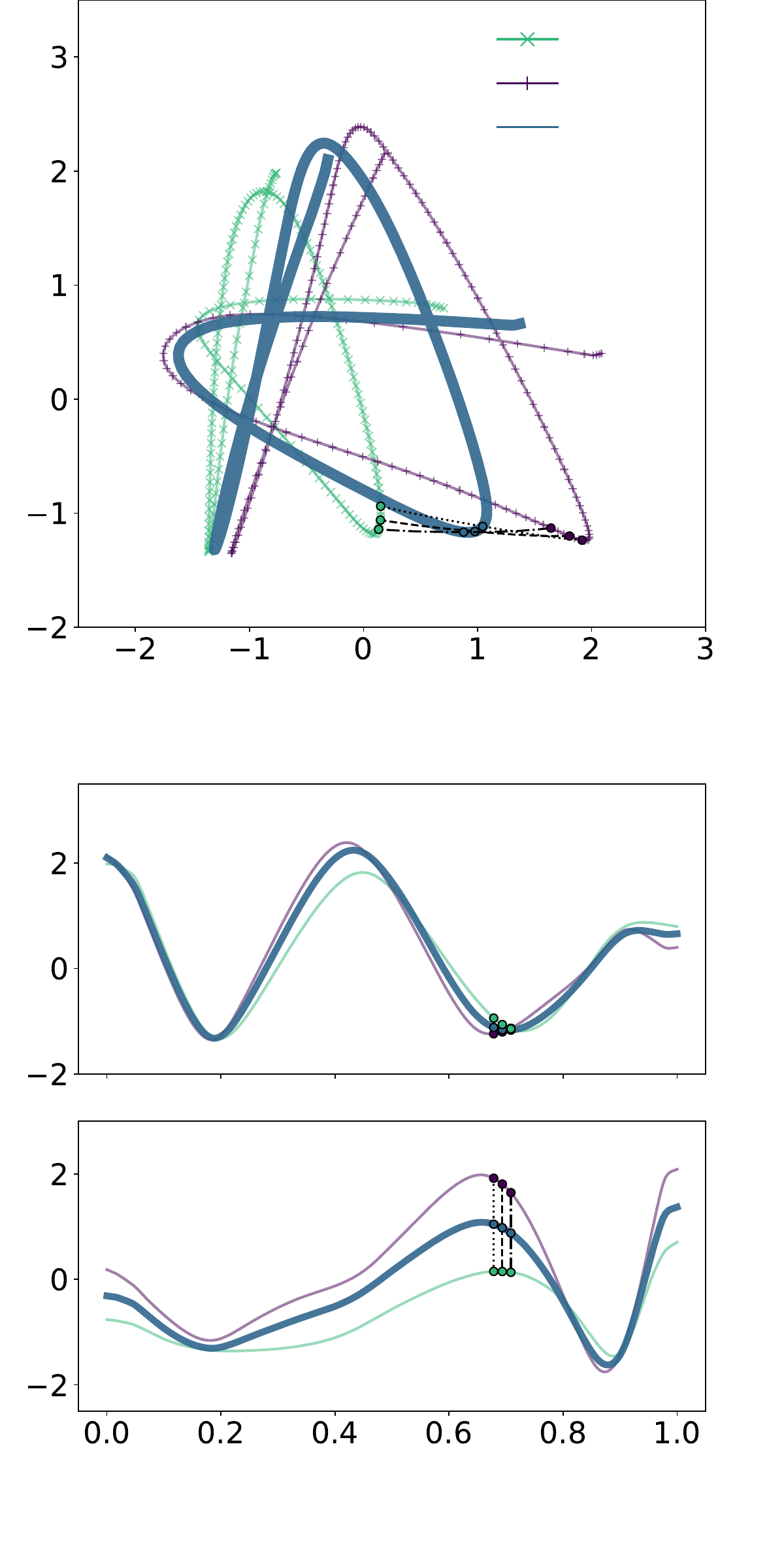}};
    \node[fill=white,minimum width=1cm,minimum height=0.1cm,inner sep=0, outer sep=0] 
      at ($(bigpic.center) + (\coff,\figxlaba)$) {\tiny X Position};
    \node[fill=white,minimum width=1cm,minimum height=0.1cm,anchor=center,rotate=90,inner sep=0, outer sep=0] 
      at ($(bigpic.west) + (0,\figylaba)$) {\tiny Y Position};
    \node[fill=white,minimum width=1cm,minimum height=0.1cm,inner sep=0, outer sep=0] 
      at ($(bigpic.south)+(\coff,\figxlabb)$) {\tiny Canonical Time};
    \node[fill=white,minimum width=1.3cm,minimum height=0.1cm,anchor=center,rotate=90,inner sep=0, outer sep=0] 
      at ($(bigpic.west) + (0,\figylabb)$) {\tiny Y Position};
    \node[fill=white,minimum width=1.3cm,minimum height=0.1cm,anchor=center,rotate=90,inner sep=0, outer sep=0] 
      at ($(bigpic.west) + (0,\figylabc)$) {\tiny X Position};
  \node[anchor=north west,font={\fontsize{5pt}{8}\selectfont}] at (\interplx,\interply) {\trajlaba};
  \node[anchor=north west,font={\fontsize{5pt}{8}\selectfont}] at (\interplx,\interply-\legendysep) {\trajlabb};
  \node[anchor=north west,font={\fontsize{5pt}{8}\selectfont}] at (\interplx,\interply-2*\legendysep) {\trajlabc};
\end{tikzpicture}%
}%
\caption{TimewarpVAE (ours)}%
\label{fig:timewarpAverage}%
\end{subfigure}%
\begin{caption}{Interpolations in latent space between canonical trajectories using various models.
For Rate Invariant Autoencoder and TimewarpVAE, we use a sixteen dimensional spatial latent space and the interpolation is constructed by 
decoding the average of the spatial latent embeddings.
The resulting average trajectory is plotted alongside
the reconstructions of the original two trajectories. 
The Rate Invariant Autoencoder can ignore parts of the canonical trajectory during training, leading to the jittering seen at the beginning and end of the canonical trajectory.
  } 
\end{caption}
\end{figure*}

Our TimewarpVAE takes the time alignment benefits of DTW
and uses 
them in a manifold learning algorithm to align the timings of similar trajectories.
Our results are shown in Fig.~\ref{fig:timewarpAverage}.
The Rate Invariant Autoencoder of \cite{Koneripalli2020} 
is similar in also
learning a latent space that separates timing and spatial factors of variation, 
with the spatial interpolations shown in Fig.~\ref{fig:rateInvariantAverage}.
Our TimewarpVAE approach has
three main contributions
that make it better suited for generating robot trajectories than a Rate Invariant AE.
The first is the 
parameterization of the generated trajectory as an arbitrarily complicated neural network function of time
rather than basing the trajectory calculation on piecewise linear functions with a pre-specified number of knots.
The second is that our approach includes a regularization term, so the model is correctly penalized for extreme time warps.
Finally, because of our time-warping regularization term, a robot can optimize the trajectory timing to account for its own joint torque and speed limitations, speeding up its execution of learned trajectories while regularizing to stay close to training timings.
An example optimized trajectory is shown in Fig~\ref{fig:trajExecution}.

\section{Related Work}

Learning a latent space of trajectories that combines the timing and spatial parameters together into a single
latent space appears in \cite{collvinent2022}, \cite{Chen2021}, and \cite{lu2019trajectory}, among others.
TimewarpVAE,
like the Rate Invariant Autoencoder of \cite{Koneripalli2020},
instead separates the timing
and spatial latent variables, giving a more efficient spatial model.
Our work is an improvement over the Rate Invariant AE in two important ways.
First, our work is not constrained to learning a piecewise linear trajectory. 
Second, we include a proper regularization term to penalize the time warping so it does not ignore parts of the template trajectory.
Comparing our results to the Rate Invariant AE shows that this timing regularization
is important to combat a degeneracy in the choice of timing of the canonical trajectory.
We explain this degeneracy in Section~\ref{sec:timingDegeneracy}
and explain how it also applies to the work of \cite{Weber2019} in the Appendix. 

Functional Data Analysis \cite{Wang2016} involves 
the study of how to interpret time-series data,
often requiring the registration of data with different timings.
The idea of warping the timings of paths appears in, for example,
DTW \cite{sakoe_dynamic_1978} and the Fr\'echet distance \cite{Frechet1906}.
Our work is also related to continuous dynamic time warping \cite{1983SymmetricTimeWarping},
which we refine for the manifold-learning setting.
The registration and averaging of trajectories is performed in \cite{Petitjean2011}, \cite{Schultz2018}, and \cite{Williams2020}.
Rather than just learning a single average trajectory, we model the full manifold of trajectories.
Time-warping is used in \cite{Chang2021} to learn ``discriminative prototypes'' of trajectories,
but not a manifold representation of all the trajectories.
A linear model to give a learned representation of registered trajectories is generated in \cite{Kneip2008},
and our work can be considered an expansion of that work, using manifold learning to allow
for nonlinear spatial variations of registered trajectories.

Time-warping has previously been combined with 
manifold learning to generate representations of 
individual frames of a trajectory. 
For example, \cite{FengZhou2012}, \cite{Trigeorgis2018}, and \cite{cho23a} align trajectories 
and learn representations of individual frames contained in the trajectories.
Connectionist Temporal Classification (CTC) can also be viewed as an algorithm for learning a labeling  
of frames of a trajectory while ignoring timing variations
\cite{Graves2006}.
Instead, our approach focuses on learning a latent vector representation that captures information about the entire trajectory.

Trajectory data
can be parameterized in many ways when presented to the learning algorithm.
For example, the trajectory could be parameterized as a dynamic movement primitive (DMP) \cite{Ijspeert2013} before learning a (linear) model of DMP parameters, as is done in \cite{Matsubara2011}.
A DMP can also be used to learn a representation of states~\cite{Chen2016} and to model trajectories;
for example,
the timing can be slowed during execution to allow a robot to ``catch up'' and correct for execution errors
\cite{Schaal2007}.
However, that work does not model timing variations during training. 
We find the Parametric DMP model is not as accurate as TimewarpVAE.
TimewarpVAE accounts for timing variations during training, 
enabling its latent variable to concentrate its modeling capacity on spatial variations of trajectories.

\section{Technical Approach}
A standard approach to learning a manifold for trajectories 
(see, for example the method proposed by \cite{Chen2012})
is to map each trajectory to a learned representation 
that includes information about timing
and spatial variations.
This type of representation learning can be performed by a beta-Variational Auto-Encoder (beta-VAE) \cite{Higgins}.
We provide a brief introduction to beta-VAE for comparison with our TimewarpVAE.

\subsection{Beta-VAE}
\begin{figure*}
  \centering
  \begin{tikzpicture}[funcBox/.style={minimum width = 2.5cm, minimum height=2cm, 
    text width=2cm, font={\fontsize{8pt}{8}\selectfont},
    align=center,
    inner sep=0,
    outer sep=0},
    varLabel/.style={font={\fontsize{8pt}{8}\selectfont}}]
    \def\vsep{1.5}
    \def\hsep{3}
    \def\vgap{0.2}
    \coordinate (xstart) at (-2*\hsep,0);
    \coordinate (spatEmb) at (0,0);
    \coordinate (dec) at (1.7*\hsep,0);
    \coordinate (end) at (2.6*\hsep,0);

    \path ($(spatEmb) + (0.7,0)$) pic[lightgray] {conv};
    \path ($(dec) + (0.9,0)$) pic[rotate=180,transform shape,lightgray] {conv};
    \node at ($(spatEmb) + (0,-0.45)$) {\small$e$};
    \node at ($(dec) + (0,-0.3)$) {\small$f$};
    \path (spatEmb) pic {myTrap={2.5}{2}};
    \path (dec) pic[rotate around={180:(dec)}] {myTrap={2.5}{2}};

    \node[funcBox] at (spatEmb) (seBox) {Encoder};
    \node[funcBox] at (dec) (decBox) {Decoder};
    \draw[rounded corners=10pt,-{latex}] (xstart) -- (seBox);
    \draw[rounded corners=10pt,-{latex}] (seBox) -- (decBox);
    \draw[rounded corners=10pt,-{latex}] (decBox) -- (end);
    \node[varLabel, anchor=south west] at ($(xstart.north east)$) {$x\in\mathbb R^{T\times n}$};
    \node[varLabel, anchor=south west] at ($(seBox.east)$) {$z\in\mathbb R^\ell$};
    \node[varLabel, anchor=north west] at ($(seBox.east)$) {$\log(\sigma^2)\in\mathbb R^\ell$};
  \node[varLabel, anchor=south west] at ($(decBox.east)$) {$\tilde x\in\mathbb R^{T\times n}$}; 
  \end{tikzpicture}
\caption{
  The architecture for Beta-VAE. 
  Beta-VAE takes in a trajectory $x$, encodes it into a latent distribution parameterized by $z$ and $\log(\sigma^2)$, and decodes to a trajectory $\tilde x$. 
}
\label{fig:autoencodeBetaVAE}
\end{figure*}
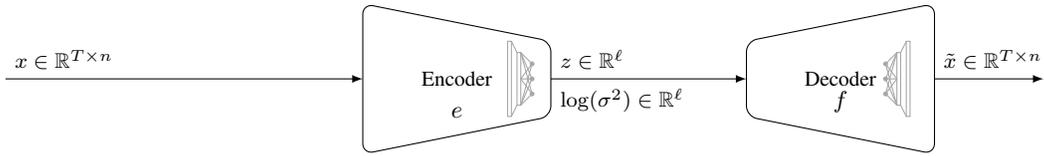

A beta-VAE 
encodes each datapoint $x$ into a learned probability distribution $q(z|x)$ in the latent space
and decodes points in the latent space into probability distributions $p(x|z)$ in the data space.
A standard formulation of beta-VAE is to parameterize the encoder distributions
as axis-aligned Gaussians $q(z|x) = \mathcal N(e(x), \sigma^2(x))$.
Thus, the encoder returns the expected latent value $z=e(x)\in\mathbb R^\ell$ for a trajectory,
along with the log of the (diagonal) covariance of the encoder noise distribution given by
$\log(\sigma^2)\in\mathbb R^\ell$.
For continuous data,
we constrain the decoder distribution to have spherical covariance with diagonal elements all equal to some constant (not-learned) value $\sigma_R^2$.
Thus, the decoder $f$ only needs to return the expected decoded trajectory $\tilde x$, given a latent value.
The beta-VAE architecture is shown in Fig.~\ref{fig:autoencodeBetaVAE}. 
Its loss function is $\mathcal L = \mathcal L_R + \mathcal L_{\mathrm{KL}}$, where
\begin{equation}
  \label{eq:betaVaeObj}
  \resizebox{\lossDefinitionScaling\width}{!}{$
              \displaystyle
  \mathcal L_R = \frac 1 {\sigma_R^2} \mathbb E_{x_i, t, \epsilon} \left[\left\| x_i(t) - 
  f\left(e(x_i)+\epsilon \right)_t \right\|^2   \right]
  $}
\end{equation}%
\begin{equation}
  \label{eq:betaVaeObj2}
  \resizebox{\lossDefinitionScaling\width}{!}{$
              \displaystyle
  \mathcal L_{\mathrm{KL}} = \beta \mathbb E_{x_i} \left[\frac {\|e(x_i)\|^2 + \sum_d (\sigma_d^2(x_i)  +\log(\sigma_d^2(x_i))}{2}\right]
  $}
\end{equation}
$\mathcal L_R$ is a reconstruction loss encouraging the model
to encode important information about the trajectories,
and 
$\mathcal L_{\mathrm{KL}}$ is a rate regularization, constraining the total amount of
information that the model is able to store about the trajectory \cite{Alemi2017}.
For clarity, we use the subscript $x_i$, to emphasize that these losses are computed as empirical expectations over each of the training trajectories.
$\sigma_d^2$ are the elements of $\sigma^2$ and
$\epsilon$ is drawn from a normal distribution with mean $0$ and diagonal covariance given by $\sigma^2$. 
$\beta$ is a regularization hyperparameter.
Since $f$ returns a full trajectory, we use the subscript $t$ to indicate indexing in to the $t$\textsuperscript{th} timestep, so it can be compared to the training pose $x_i(t)$.

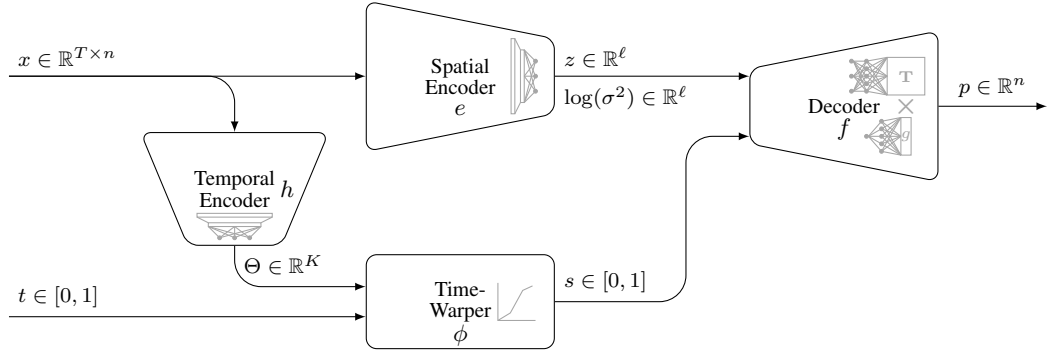
\begin{figure*}
  \centering
  \begin{tikzpicture}[funcBox/.style={minimum width = 2.5cm, minimum height=1.25cm, 
    text width=2cm, font={\fontsize{8pt}{8}\selectfont},
    align=center},
    varLabel/.style={font={\fontsize{8pt}{8}\selectfont}}]
    \def\vsep{1.5}
    \def\hsep{3}
    \def\vgap{0.2}
    \coordinate (xstart) at (-2*\hsep,\vsep);
    \coordinate (tstart) at (-2*\hsep,-\vsep-\vgap) (tstart);
    \coordinate (tempEmb) at (-\hsep,0);
    \coordinate (spatEmb) at (0,\vsep);
    \coordinate (timeWarp) at (0,-\vsep);
    \coordinate (twDecBend) at (\hsep, \vsep-4*\vgap);
    \coordinate (dec) at (1.7*\hsep,\vsep-2*\vgap);
    \coordinate (end) at (2.6*\hsep,\vsep-2*\vgap);
    \node[funcBox,minimum height=1.5cm] at (tempEmb) (teBox) {Temporal\\ Encoder};
    \node[funcBox] at (spatEmb) (seBox) {Spatial\\ Encoder};
    \node[funcBox,rounded corners,draw] at (timeWarp) (twBox) {Time-\\ Warper};
    \node[funcBox] at (dec) (decBox) {Decoder};
    \path (spatEmb) pic {myTrap={2.5}{2}};
    \path (dec) pic[rotate around={180:(dec)}] {myTrap={2.5}{2}};
    \path (tempEmb) pic[rotate around={270:(tempEmb)}] {myTrap={1.5}{2.5}};
    \path ($(spatEmb) + (0.7,0)$) pic[lightgray] {conv};
    \path ($(dec) + (0.1,0)$) pic[lightgray] {factorizedDecoderImg};
    \path ($(timeWarp) + (0.75,0)$) pic[lightgray] {timeWarperImg};
    \path ($(tempEmb) - (0,0.35)$) pic[rotate=-90,transform shape,lightgray] {conv};
    \node at ($(spatEmb) + (0,-0.45)$) {\small$e$};
    \node at ($(dec) + (0,-0.3)$) {\small$f$};
    \node at ($(tempEmb) + (0.7,0)$) {\small$h$};
    \node at ($(timeWarp) + (0,-0.45)$) {\small$\phi$};
    \draw[rounded corners=10pt,-{latex}] (xstart) -| (teBox);
    \draw[rounded corners=10pt,-{latex}] (teBox)  |- ($(twBox.west)+(0,\vgap)$);
    \draw[rounded corners=10pt,-{latex}] (xstart) -- (seBox);
    \draw[rounded corners=10pt,-{latex}] (tstart) -- ($(twBox.west) - (0,\vgap)$);
    \draw[rounded corners=10pt,-{latex}] (twBox) -| (twDecBend) -- ($(decBox.west) - (0,2*\vgap)$);
    \draw[rounded corners=5pt,-{latex}] (seBox) -- ($(decBox.west) + (0,2*\vgap)$);
    \draw[rounded corners=10pt,-{latex}] (decBox) -- (end);
    \node[varLabel, anchor=south west] at ($(xstart.north east)$) {$x\in\mathbb R^{T\times n}$};
    \node[varLabel, anchor=south west] at ($(tstart.north east)$) {$t\in[0,1]$};
    \node[varLabel, anchor=north west] at ($(teBox.south)$) {$\Theta\in\mathbb R^K$};
    \node[varLabel, anchor=south west] at ($(twBox.east)$) {$s\in[0,1]$};
    \node[varLabel, anchor=south west] at ($(seBox.east)$) {$z\in\mathbb R^\ell$};
    \node[varLabel, anchor=north west] at ($(seBox.east)$) {$\log(\sigma^2)\in\mathbb R^\ell$};
  \node[varLabel, anchor=south] at ($0.5*(decBox.east) + 0.5*(end.west)$) {$p\in\mathbb R^n$}; \end{tikzpicture}%
\caption{
  The architecture for TimewarpVAE. 
  TimewarpVAE takes in a full trajectory $z$ and a timestamp $t$ and reconstructs the position $p$ of the trajectory at that timestamp. TimewarpVAE separately encodes the timing of the trajectory into $\Theta$ and encodes the spatial information into a latent distribution parameterized by $z$ and $\log(\sigma^2)$.}
\label{fig:autoencode}%
\end{figure*}

\subsection{TimewarpVAE}
A standard approach to learning a manifold for trajectories 
(see, for example, the method proposed by \cite{Chen2012})
is to map each trajectory to a learned representation 
that includes information about timing and spatial variations.
This type of representation learning can be performed by a beta-VAE \cite{Higgins}.
TimewarpVAE is based on beta-VAE,
with the goal of 
separating latent variables for spatial and timing factors
to make the models more useful for robot execution.
To separate the spatial and temporal variations in trajectories,
TimewarpVAE 
contains two modules not present in beta-VAE: a temporal encoder and a time-warper.
The decoder now takes in information from both the spatial encoder and the time-warper.
Fig.~\ref{fig:autoencode} shows the architecture of TimewarpVAE.
It takes in two inputs, the training trajectory $x$
and a desired reconstruction time $t$.
Like in beta-VAE, the spatial encoder maps the trajectory $x$ to its latent value distribution
parameterized by $z$ and $\log(\sigma^2)$.
The temporal encoder computes time-warping parameters $\Theta$,
and the time-warper (defined by the parameters $\Theta$) now acts on $t$ 
to warp it to a ``canonical time'' $s$.
The decoder takes the canonical time $s$ and the spatial latent vector
and returns
the position of the canonical trajectory for that latent vector at the canonical time $s$.
These modules are further explained in Section~\ref{sec:nnTimeWarper}.
These functions are jointly trained so the decoded position 
is a good reconstruction of the position of trajectory $x$ at timestep $t$,
while at the same time minimizing the information that we allow the autoencoder to 
store about the trajectory. 

Specifically, the minimization objective for TimewarpVAE, denoted $\mathcal L$, is the sum of the reconstruction cost $\mathcal L_R$, beta-VAE's KL divergence loss $\mathcal L_{\mathrm{KL}}$, and a new time-warping regularization $\mathcal L_\phi$, which we explain further in Section~\ref{sec:timingDegeneracy}.
  $\mathcal L = \mathcal L_R + \mathcal L_{\mathrm{KL}} + \mathcal L_{\phi}$,
where
\begin{equation}
  \label{eq:TimewarpVAEObjective}
  \resizebox{\lossDefinitionScaling\width}{!}{$
              \displaystyle
  \mathcal L_R = \frac 1 {\sigma_R^2} \mathbb E_{x_i, t, \epsilon} \left[\left\| x_i(t) - 
  f\left(
  \sum_{j=1}^{k} h(x_i)_j \psi_j(t)
  ,\, e(x_i)
  +\epsilon \right) 
  \right\|^2   \right]
  $}
\end{equation}%
\begin{equation}
  \resizebox{\lossDefinitionScaling\width}{!}{$
              \displaystyle
  \mathcal L_{\mathrm{KL}} = \beta \mathbb E_{x_i} \left[\frac {\|e(x_i)\|^2 + \sum_d (\sigma_d^2(x_i)  +\log(\sigma_d^2(x_i))}{2}\right]
  $}
\end{equation}%
\begin{equation}
  \resizebox{\lossDefinitionScaling\width}{!}{$
              \displaystyle
  \mathcal L_{\phi} = \lambda \mathbb E_{x_i} \left[\frac 1 K \sum_{j=1}^K (h(x_i)_j-1)\log(h(x_i)_j)\right]
  $}
\end{equation}%

For clarity, we again use the subscript $x_i$ to emphasize that these losses are computed as empirical expectations over each training trajectory $x_i$.
$e$, $\sigma^2$ (and its elements $\sigma_d^2$), $\epsilon$, and $\sigma_R^2$ are all defined in the same way as for beta-VAE in Eqs.~\ref{eq:betaVaeObj} and~\ref{eq:betaVaeObj2}.
$f$ is the decoder, now taking in a canonical timestamp along with the latent value.
$h$ is the temporal encoder, so that $h(x_i)_j$ is the $j$th output neuron (out of $K$ total) of the temporal encoder applied to the $i$th trajectory.
The $\psi_j$ are the time-warping basis functions, explained in Sec.~\ref{sec:nnTimeWarper} and defined in Equation~\ref{eq:psi}.
$\beta$ is a regularization hyperparameter for beta-VAEs.
$\lambda$ is a regularization hyperparameter for our time-warping functions, which we set to 0.05 in our experiments.
Since $\lambda$ penalizes the time-warping in the model, a large $\lambda \to \infty$ would
drive time-warping to be the identity function, and $\lambda=0$ could allow the model to learn severe time-warps.
We explain the neural network implementation of all of these functions in the next section.

The benefits of this algorithm are as follows:
  it learns a low-dimensional representation of spatial variations in trajectories;
  it can be implemented as a neural network and trained using backpropagation;
  and it can accommodate nonlinear timing differences in the training trajectories.
Additionally, new trajectories can be generated using a latent value $z$
and canonical timestamps $s$ ranging from 0 to 1
without using the time-warper or the temporal encoder,
which we do in our empirical evaluations, calling these generated trajectories ``canonical'' or ``template'' trajectories.
These trajectories outperform baseline trajectories qualitatively in Fig.~\ref{fig:timewarpAverage} and quantitatively in our results section.

\section{Neural Network Formulation}
\label{sec:nnTimeWarper}
This section explains how to write the spatial encoding function, the temporal encoding function, the time-warping function, and the decoding function as differentiable
functions.
The time-warping function is a differentiable function with no learnable parameters since the time-warping is entirely defined by the input parameters $\Theta$.
The other three modules have learnable parameters, which we learn through backprop.

\paragraph{Architecture for the time-warper.}
The time-warper takes in
a training timestamp $t$ for a particular trajectory and maps it monotonically to a canonical timestamp $s$.
$\phi$ is a piecewise linear function of $t$ with equally spaced knots and $K$ linear segments.
The slopes of those segments are labeled by $\Theta_j$ for $1 \le j \le K$.
Different vector $\Theta \in \mathbb R^K$ choices give different time-warping functions.
In order for $\Theta$ to yield a valid time-warping function mapping $[0,1]$ to $[0,1]$,
the $\Theta_j$ should be positive and average to $1$.
These $\Theta$ values are generated by the temporal encoding function discussed in the next paragraph.
Given some vector $\Theta$, corresponding to the slope of each segment,
the time-warper $\phi$ is 
given by 
$\phi(t) = \sum_{j=1}^{K} \Theta_j \psi_j(t)$
where the $\psi_j$ are defined as follows 
and do not need to be learned:
\begin{equation}
  \label{eq:psi}
  \psi_j(t) = \min\Big\{\max\big\{t - (j-1)/K, 0\big\}, 1/K\Big\}
\end{equation}
A visualization of these basis functions $\psi_j$ is presented in the Appendix.
We use the specific parameterization of $\Theta_j$ described in the next paragraph
to ensure that 
our time-warping function is a bijection from $[0,1]$ to $[0,1]$.

\paragraph{Neural network architecture for the temporal encoder.} 
A neural network $h : \mathbb R^{n \times T} \to \mathbb R^K$
computes a different vector $\Theta$ for each training trajectory $x$.
To ensure the elements of $\Theta$ are positive and average to $1$,
the softmax function is applied to 
the last layer of the temporal encoder, and the result is scaled by $K$.
This transformation sends the values $\theta$ to 
${\Theta_j = \mathrm{Softmax}\left(\theta\right)_j} K$
for $j$ from $1$ to $K$,
ensuring that 
the average of the output neurons $\Theta_j$ is $1$ as desired.
By contrast, \cite{Koneripalli2020} square the last layer and then normalize.

\paragraph{Neural network architecture for the spatial encoder.}
Given a trajectory $x$ evenly sampled at $T$ different timesteps $t_j$ between $0$ and $1$, the $T\times n$ matrix of these evenly sampled positions $x(t_j)$  is written as $x\in \mathbb R^{n \times T}$.
In the neural network architecture used in our experiments, one-dimensional convolutions are applied over the time dimension, treating the $n$ spatial dimensions as input channels.
This is followed by different fully connected layers for $e$ and for $\log(\sigma)$.
However, any neural network architecture, such as a Transformer \cite{transformer} or Recurrent Neural Network \cite{lstm}, could be used in the spatial encoder module of a TimewarpVAE.

\paragraph{Neural network architecture for the decoder}
Any architecture that takes in a time $s$ and a latent vector $z$ and returns a position $p$ could be used for the decoder $f$.
Our experiments use a modular decomposition of $f$, 
with $f(s,z)$ decomposed as the product of a matrix and a vector: $f(s,z)=\mathbf T(z)g(s)$.
In this formulation, the matrix $\mathbf T(z)$ is a function of the latent vector $z$, and the vector $g(s)$ is a function of the (retimed) timestep $s$. 
If each point in the training trajectory has dimension $n$,
and for some hyperparameter $m$ for our architecture,
the matrix $\mathbf T(z)$ will have shape $n\times m$, and the vector $g(s)$ will be of length $m$.
The $nm$ elements of $\mathbf T(z)$ are the (reshaped) output of a sequence of fully connected layers
taking the vector $z$ as an input.
The $m$ elements of $g(s)$ are computed as the output of a sequence of fully connected layers
taking the scalar $s$ as an input.
Because the scalar $s$ will lie in the range $[0,1]$,
it is useful to customize the initialization
of the weights in the first hidden layer of $g(s)$.
Details of the custom initialization are provided in the Appendix. 
This chosen architecture is useful for easily creating the ``NoNonlinearity'' ablation experiment in Section~\ref{sec:ablations}.
With this architecture, if the hidden layers in $\mathbf T(z)$ are removed, then it a linear function of $z$, and so
then entire decoder becomes linear with respect to $z$, so the generated position $p$ will be a linear combination
of possible positions at timestep $s$.

\subsection{Regularization of Time-Warping Function}
\label{sec:timingDegeneracy}
The choice of timing for the canonical trajectories adds a degeneracy to the solution.\footnote{
 This is similar to the degeneracy noted 
 in the Appendix for continuous dynamic time warping
\cite{1983SymmetricTimeWarping}, and, as noted in the Appendix, was not properly analyzed in \cite{Weber2019}.
}
Without our regularization, it is possible for other methods, like Rate Invariant AE,
to warp so severely that they can ignore parts of the canonical trajectory.
This is a problem if we generate robot motions from the canonical trajectory since it can lead to jittering motions,
as seen at the beginning and end of the trajectory in Fig.~\ref{fig:rateInvariantAverage}.

  We propose a regularization penalty on the time-warper $\phi$ 
  to choose among the degenerate solutions.
  The proposed penalty is on 
  $\int_0^1 \left(\phi'(t)-1\right) \log\left( \phi'(t)\right)\,dt$.
That regularization contains the function $g(x) = (x-1)\log(x)$ applied to the slope $\phi'(t)$, integrated over each point $t$.
That function $g(x)$ is concave-up for $x>0$ and takes on a minimum at $x=1$, encouraging the slope of $\phi$ to be near $1$.
This formulation 
has the symmetric property that regularizing $\phi(t)$
gives the exact same regularization as regularizing $\phi^{-1}(s)$. This symmetry is proven in the Appendix.
For each trajectory $x_i$ our time-warper $\phi$ is stepwise linear with $K$ equally-sized segments
with slopes $\Theta_1=h(x_i)_1,\ldots,\Theta_k=h(x_i)_K$.
Thus, the regularization integral for the time-warper associated with $x_i$ is
\begin{equation}
  \mathcal L_\phi(x_i) = 
  \frac 1 K \sum_{j=1}^K (h(x_i)_j-1)\log(h(x_i)_j)
\end{equation}

\section{Experiments}
\begin{figure}
\centering
  \begin{subcaptionbox}{The demonstration system 
  \label{fig:recordingSetup}
    }[0.23\textwidth]{
  \centering
    \resizebox{4cm}{!}{
    \begin{tikzpicture}
    \node at (0,0) {\includegraphics[trim=400 100 500 120, clip,width=6cm]{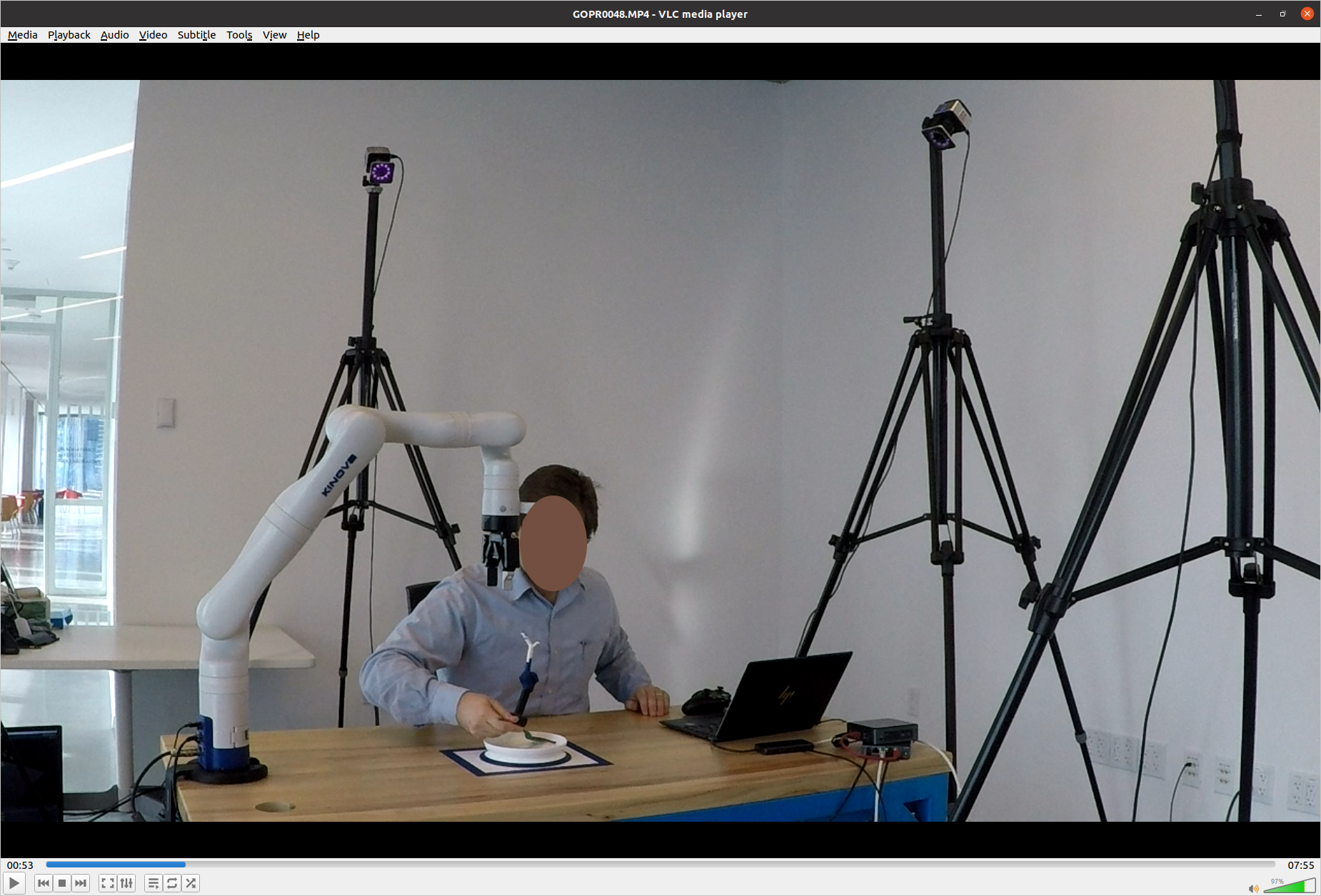}};
    \node[rectangle, fill=white] at (0,2) (vcamtext) {Vicon Cameras};
    \draw[-{latex},thick] (vcamtext) -> (2.5,2.7);
    \draw[-{latex},thick] (vcamtext) -> (-2.0,2.5);
    \node[rectangle, fill=white] at (1.5,-1) (vmarktext) {Vicon Markers};
    \draw[-{latex},thick] (vmarktext) -> (-0.7,-1.8);
    \node[rectangle, fill=white] at (1.5,-2) (forktext) {Fork};
    \draw[-{latex},thick] (forktext) -> (-0.8,-2.5);
   \end{tikzpicture}
 }
}
\end{subcaptionbox}
\centering
  \begin{subcaptionbox}{Six demonstrations\footnotemark \label{fig:demonstrationForkTrajectories}}
    [0.23\textwidth]{
\centering
    \resizebox{4cm}{!}{
  \begin{tikzpicture}
  \def \vsep{2.5}
  \def \hsep{3.7}
    \useasboundingbox (-0.75*\hsep,-0.5*\vsep) rectangle (1.75*\hsep/4, 1.4*\vsep);
  \node at ($(0,0)$) {\includegraphics[trim=150 230 150 200,clip,width=3.5cm]{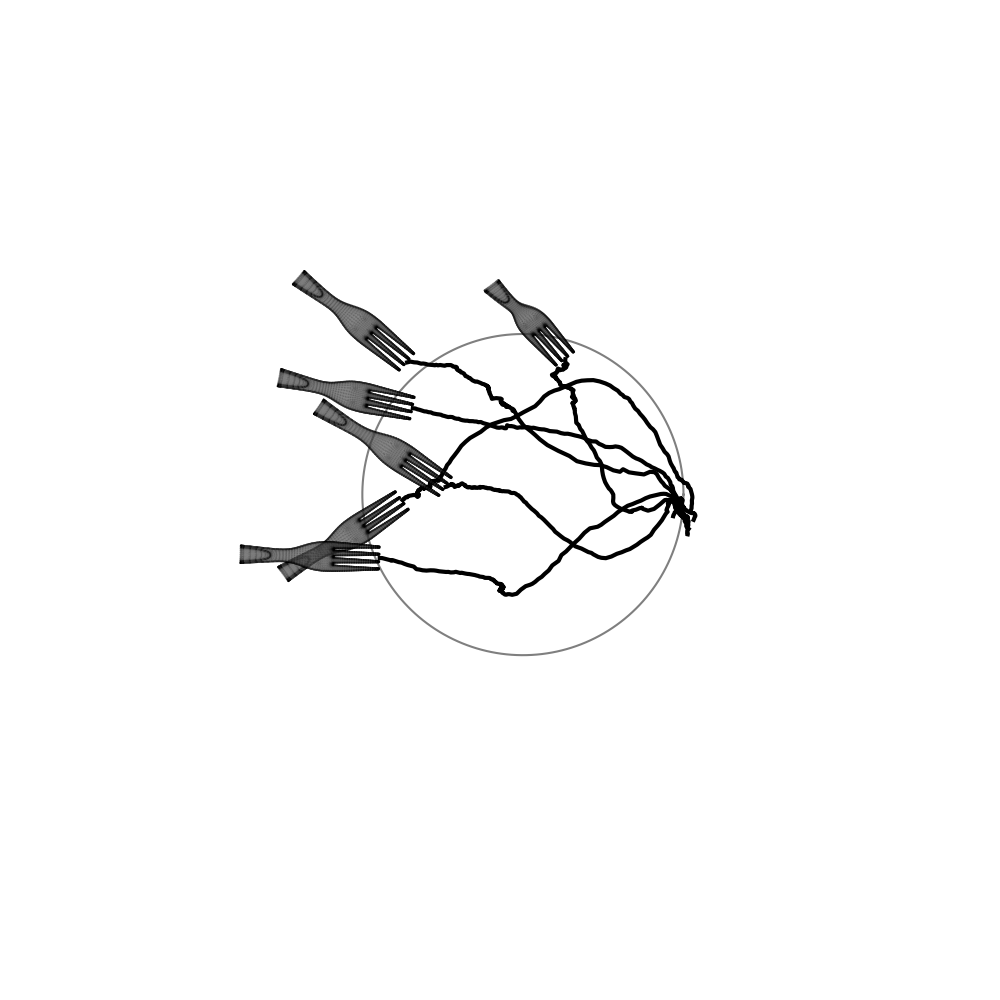}};
  \node at ($(0,\vsep)$) {\includegraphics[trim=150 190 150 250,clip,width=3.5cm]{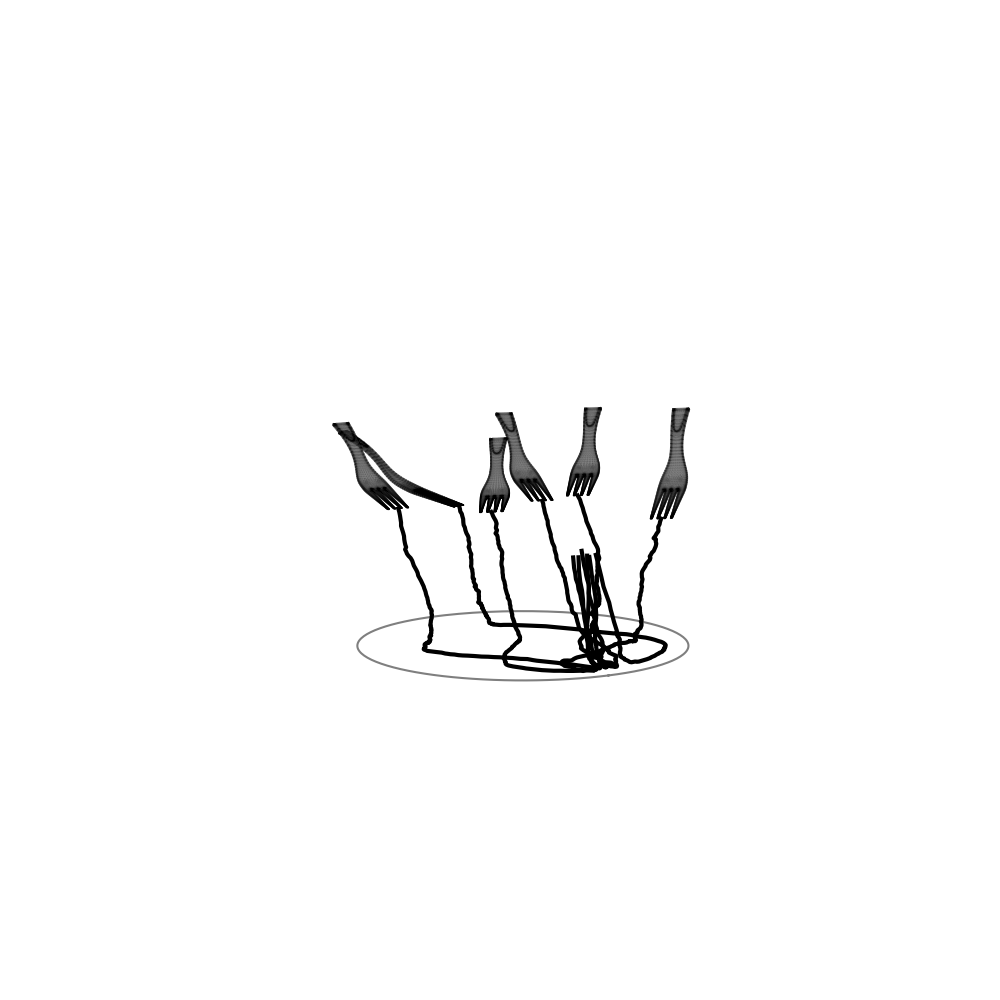}};
    \node[rotate=90] at ($(-2.4,0)$) {Top view};
    \node[rotate=90] at ($(-2.4,\vsep)$) {Angled view};
\end{tikzpicture}
}}
\end{subcaptionbox}
  \caption{We collect trajectory recordings of the position and orientation of a fork 
  while it is used to pick a small piece of yarn off a plate with steep sides. 
  Example trajectories are presented from two angles, showing the initial orientation of the fork and the position of the tip of the fork over time.}
\end{figure}
Experiments are performed on two datasets, one containing handwriting gestures made in the air while holding a Wii remote \cite{chen_6dmg_2012}, 
and one that we collect ourselves, containing motions that pick up yarn off a plate with a fork, mimicking food acquisition.
The same model architecture is used for both experiments,
with hyperparameters given in the Appendix. 
Additionally, during training, we perform data augmentation by randomly perturbing the timesteps used when sampling the trajectory $x$,
using linear interpolation to compute sampled positions.
Our specific data-augmentation implementation is described in the Appendix. 
This data augmentation decreases training performance but greatly improves test performance, as shown in the ablation studies.
\footnotetext{Fork meshes in 3D trajectory plots were downloaded and modified from https://www.turbosquid.com/3d-models/metal-fork-3d-model/362158 and are used under the TurboSquid 3D Model License}

\subsection{Fork Trajectory Dataset}
345 fork trajectories are recorded using
the Vicon tracking system shown in
Fig.~\ref{fig:recordingSetup}. Reflective markers are rigidly attached to a plastic fork, and the trajectory of the fork is recorded using Vicon cameras.
A six-centimeter length of yarn is placed on the plate in an arbitrary location and orientation. 
It is then picked up right-handed by scraping it to the left and using the side of the plate as a static tool to push it onto the fork.
Demonstrations were intentionally collected with three different timings,
where in some trajectories the approach to the plate was intentionally much faster than the retreat from the plate, in some trajectories the approach was intentionally much slower than the retreat from the plate, and in the remaining trajectories the approach and retreat are approximately the same speed.
The dataset was split into 240 training trajectories and 105 test trajectories.
Examples of six recorded trajectories, along with visualizations of the starting pose of the fork for those trajectories,
are presented in Fig.~\ref{fig:demonstrationForkTrajectories}.
Trajectories are truncated to start when the fork tip passes within 10cm of the table and to end
again when the fork passes above 10cm.
All trajectories were subsampled to 200 equally-spaced time points, using linear interpolation as needed.
We express the data as the $x,y,z$ position of the tip of the fork and the $rw,rx,ry,rz$ quaternion orientation of the fork,
giving a dataset of dimension $n=7$ at each datapoint.
The data is preprocessed to choose a consistent sign of the quaternion representations so they are all near each other in $\mathbb R^4$.
The data is mean-centered by subtracting the average $x,y,z,rw,rx,ry,rz$ training values,
and the $x,y,z$ values are divided by a normalizing factor
so that their combined variance $\mathbb E[x^2 + y^2 + z^2]$ is 
$3$ on the training set.
Likewise, the $rw,rx,ry,rz$ values are multiplied by a scaling factor of $0.08m$,
(chosen as a length scale associated with the size of the fork),
before dividing by the same normalizing factor,
to bring all the dimensions into the same range.

\subsection{Handwriting Gestures Dataset}
The
air-handwriting dataset collected by~\cite{chen_6dmg_2012} is used for the handwriting experiment.
The drawn letters are projected onto xy coordinates. 
A training set of 125 random examples of the letter ``A'' are drawn from the air-handwriting dataset
, and the remaining 125 random examples of the letter ``A'' are the test set.
All trajectories were subsampled to 200 equally-spaced time points, using linear interpolation as needed.
The data are mean-centered so that the average position over the whole training dataset is the origin,
and $x$ and $y$ are scaled together by the same constant 
so that their combined variance $\mathbb E[x^2 + y^2]$ is 
$2$ on the training set.
Example training trajectories are in the Appendix.

\begin{table}
  \scriptsize
  \begin{tabular}{lrrrr}
   \toprule
    Architecture & Beta & Rate & Training A-RMSE & Test A-RMSE \\
    & & & ($\pm 3\sigma$) & ($\pm 3\sigma$)\\
    \midrule
  TimewarpVAE & 0.01 & 3.716 & 0.187 $\pm$ 0.003 & \textbf{0.233 $\pm$ 0.003} \\
 & 0.1 & 3.227 & \textbf{0.185 $\pm$ 0.007} & 0.234 $\pm$ 0.008 \\
RateInvariantAE & 0.01 & 4.095 & 0.260 $\pm$ 0.130 & 0.316 $\pm$ 0.188\\
 & 0.1 & 3.280 & 0.285 $\pm$ 0.154 & 0.325 $\pm$ 0.132\\
beta-VAE & 0.01 & 4.759 & 0.291 $\pm$ 0.005 & 0.343 $\pm$ 0.016\\
 & 0.1 & 3.670 & 0.293 $\pm$ 0.007 & 0.342 $\pm$ 0.011\\
NoTimewarp & 0.01 & 3.924 & 0.264 $\pm$ 0.007 & 0.360 $\pm$ 0.017\\
 & 0.1 & 3.508 & 0.265 $\pm$ 0.006 & 0.354 $\pm$ 0.014\\\
    \bottomrule
  \end{tabular}
  \centering
  \caption{Performance results for 3-dimensional models of fork trajectories. Our TimewarpVAE significantly outperforms beta-VAE and the ablation of TimewarpVAE without the time-warper.}
  \label{tab:forkPerformance}
\end{table}

\subsection{Model Performance Measures}
The performance measures include three important values: the training reconstruction error, 
the test reconstruction error,
and the model rate.
To measure spatial variation of trajectories,
reconstruction errors are computed
by first performing symmetric DTW to align the reconstructed trajectory
to the original trajectory.
We then compute the Euclidean mean squared error
between each point in the original trajectory and every point it is paired with.
After that, we calculate the average of all those errors over all the timesteps in the original trajectory
before taking the square root to compute our aligned root mean squared error (A-RMSE).
In the framework of Rate-Distortion theory, these errors are distortion terms.
The model rate measures the information bottleneck imposed by the model and is given by the KL divergence term in the beta-VAE loss function.
It's important to check the rate of the model
since arbitrarily complex models can get perfect reconstruction error if they have a large enough rate \cite{alemi_fixing_2018}.
However, among trained models with similar rates, it is fair to say that the model with lower
distortion is a better model.
Model rate is reported in bits.

\begin{figure}
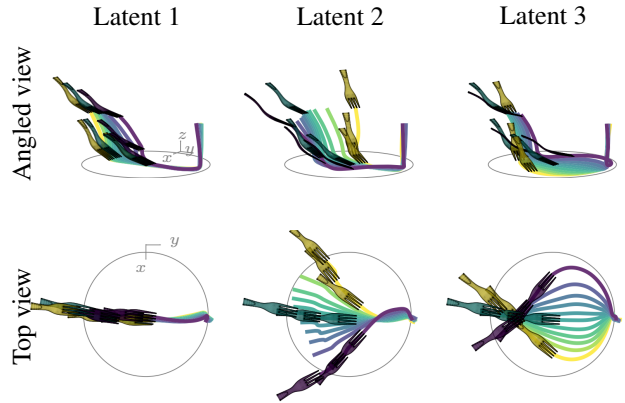

\centering
  \begin{tikzpicture}
  \def \vsep{2.7}
  \def \hsep{2.7}
    \useasboundingbox (-0.55*\hsep,-\vsep/2) rectangle (2*\hsep+2*\hsep/4, 1.75*\vsep);
\foreach \latent in {0,...,2} {%
    \node at ($(\hsep*\latent,0)$) (bswep\latent) {\includegraphics[trim=150 200 150 200,clip,width=3cm]{latent\latent-elev90-azim0}};
    \node at ($(\hsep*\latent,\vsep)$) (aswep\latent) {\includegraphics[trim=150 230 150 250,clip,width=3cm]{latent\latent-elev10-azim30}};
} 
    \node at ($(bswep0) + (0.05,1.6*0.40)$) {\tiny\color{gray}$x$};
    \node at ($(bswep0) + (0.5,1.8*0.55)$) {\tiny\color{gray}$y$};
    \node at ($(aswep0) + (0.4,-0.6)$) {\tiny\color{gray}$x$};
    \node at ($(aswep0) + (0.7,-0.5)$) {\tiny\color{gray}$y$};
    \node at ($(aswep0) + (0.6,-0.3)$) {\tiny\color{gray}$z$};
\foreach \latent in {1,...,3} {%
  \node at ($(\hsep*\latent-\hsep, \vsep + 1.3)$) {Latent \latent};
}
    \node[rotate=90] at ($(-1.5,0)$) {Top view};
    \node[rotate=90] at ($(-1.5,\vsep)$) {Angled view};
\end{tikzpicture}
  \caption{
  The paths of the fork tip are plotted over time for TimewarpVAE trajectories using different latent vectors. The orientation of the fork is shown at three different timesteps in a color matching the associated path.
  Each latent dimension has an interpretable effect.
  The first latent determines the fork's initial $y$ position, the second latent determines the fork's initial $x$ position, and the third latent determines how the fork curves during trajectory execution.}
  \label{fig:latentSweep}
\end{figure}

\subsection{Fork Model Results}
Models are trained with a latent dimension of three on the fork dataset. TimewarpVAE is compared to beta-VAE, a VAE version of Rate Invariant AE, and an ablation of TimewarpVAE called ``NoTimewarp'' that sets the time-warping module to the identity function.
The latent sweep of a TimewarpVAE trained with $\beta=1$ is shown in Fig.~\ref{fig:latentSweep}, showing its interpretable latent space.
Table~\ref{tab:forkPerformance} shows performance measures, where
we compute and summarize five trials for each model type for various hyperparameters $\beta$.
TimewarpVAE outperforms the baseline methods.

\subsection{Air Handwriting Results}
\label{sec:ablations}
TimewarpVAE is compared to baseline methods trained on the same training and test splits.
All models are trained with a batch size of 64 for 20,000 epochs, using the Adam optimizer and a learning rate 0.001.
We train each model five times for different choices of beta,
and we plot the mean and one standard error above and below the mean.
To give context to these results, we also show results for parametric dynamic movement primitives (Parametric DMPs)~\cite{Matsubara2011}
and PCA results, which do not have associated rate values.
TimewarpVAE significantly outperforms Parametric DMPs, PCA, and beta-VAE, with the greatest differences for smaller latent dimensions.
TimewarpVAE shows comparable performance to RateInvariant on training error and consistently outperforms RateInvariant in test error.

\begin{figure*}
\centering%
\begin{subfigure}[b]{0.3\textwidth}%
\centering%
\begin{tikzpicture}
  \useasboundingbox (-2.2,-1.6) rectangle (2.2,2);
  \node at (0,0) {\includegraphics[width=4cm]{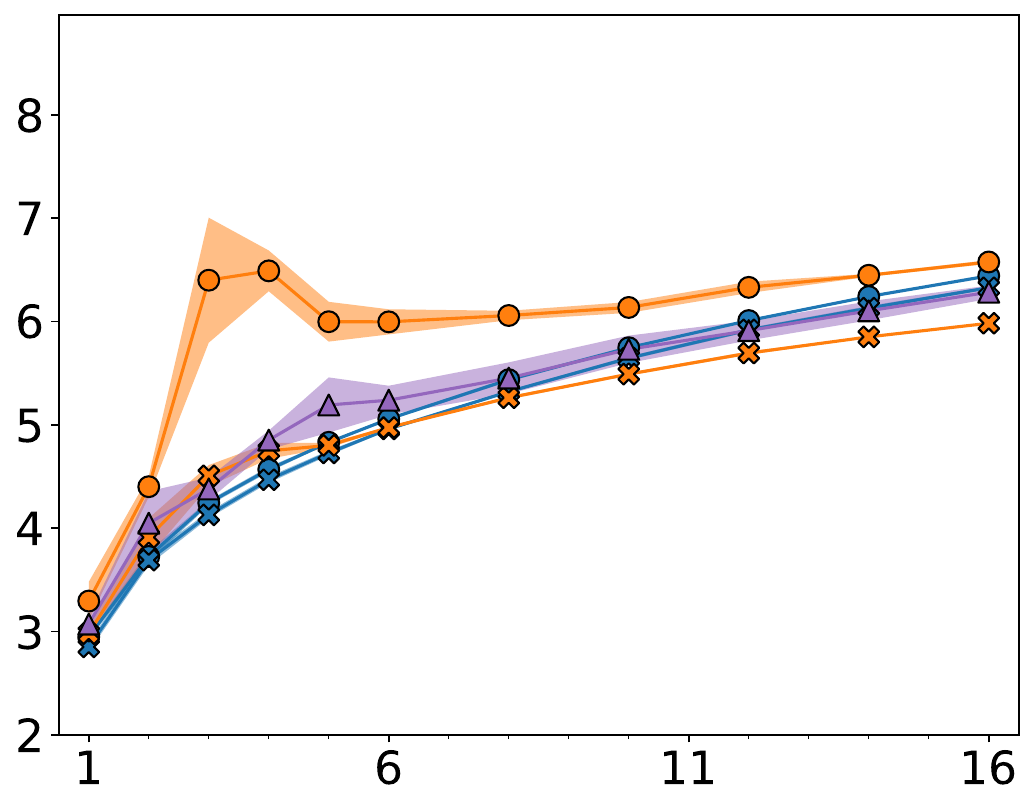}};
  \node at (0.15,-1.6) {\tiny Latent dimension};
  \node[anchor=center,rotate=90] at (-2.1,0) {\tiny Bits};
\end{tikzpicture}
  \caption{Rate}
\end{subfigure}
\begin{subfigure}[b]{0.3\textwidth}%
\centering%
\begin{tikzpicture}
  \useasboundingbox (-2.2,-1.6) rectangle (2.2,2);
  \node at (0,0) {\includegraphics[width=4cm]{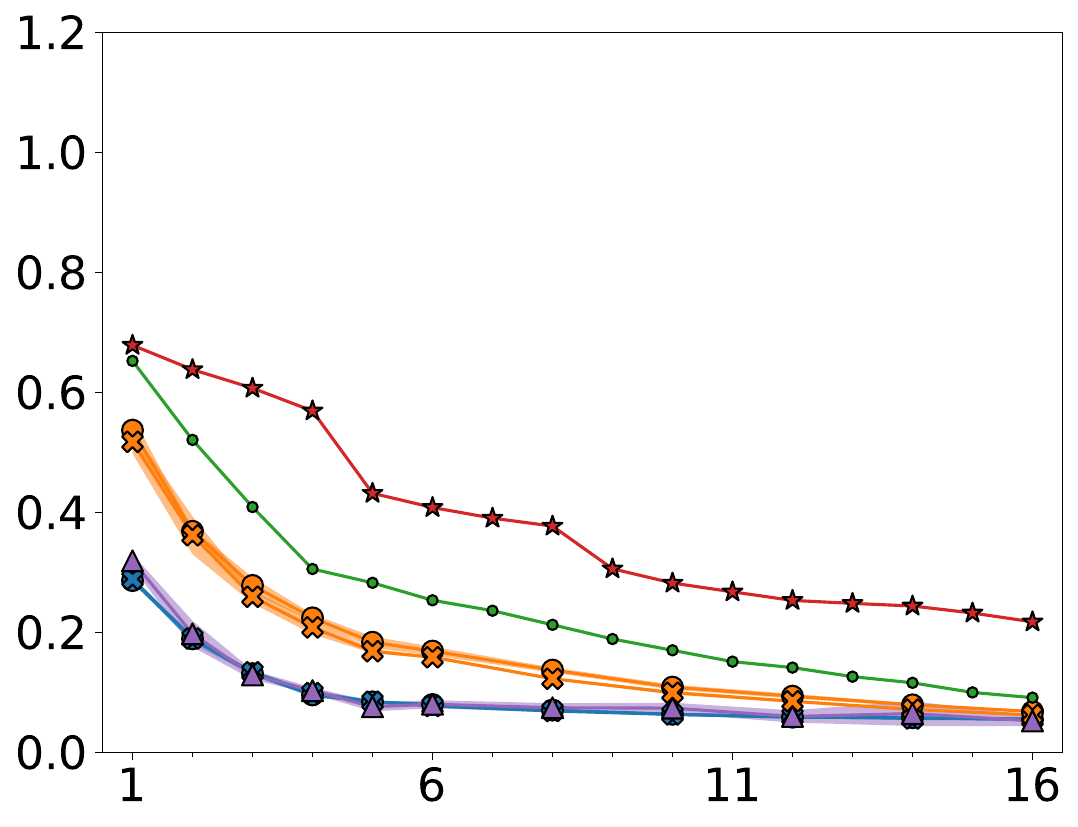}};
  \node at (0.15,-1.6) {\tiny Latent dimension};
  \node[anchor=center,rotate=90] at (-2.1,0) {\tiny A-RMSE};
\end{tikzpicture}
  \caption{Train Distortion}
\end{subfigure}
\begin{subfigure}[b]{0.3\textwidth}%
\centering%
\begin{tikzpicture}
  \useasboundingbox (-2.2,-1.6) rectangle (2.2,2);
  \node at (0,0) {\includegraphics[width=4cm]{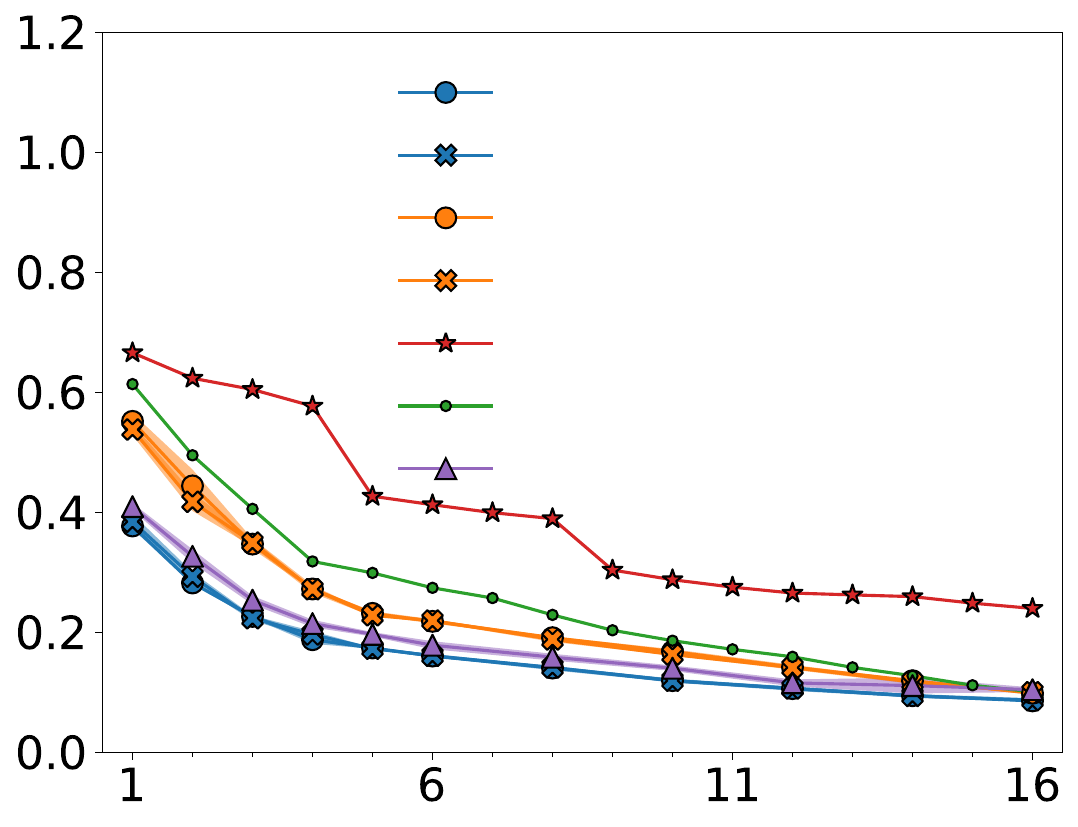}};
  \node at (0.15,-1.6) {\tiny Latent dimension};
  \node[anchor=center,rotate=90] at (-2.1,0) {\tiny A-RMSE};
  \def\lx{-0.30}
  \def\ly{1.4}
  \node[anchor=north west,font={\fontsize{5pt}{8}\selectfont}] at (\lx,\ly) {$\beta$=0.001};
  \node[anchor=north west,font={\fontsize{5pt}{8}\selectfont}] at (\lx,\ly-\lsep) {$\beta$=0.01};
  \node[anchor=north west,font={\fontsize{5pt}{8}\selectfont}] at (\lx,\ly-2*\lsep) {$\beta$=0.001};
  \node[anchor=north west,font={\fontsize{5pt}{8}\selectfont}] at (\lx,\ly-3*\lsep) {$\beta$=0.01};
  \node[anchor=north west,font={\fontsize{5pt}{8}\selectfont}] at (\lx+0.9,\ly-0.5*\lsep) (base){TimewarpVAE};
  \node[anchor=north east,font={\fontsize{5pt}{8}\selectfont}] at (base.east){(ours)};
  \node[anchor=north west,font={\fontsize{5pt}{8}\selectfont}] at (\lx+0.9,\ly-2.5*\lsep){Beta-VAE};
  \node[anchor=north west,font={\fontsize{5pt}{8}\selectfont}] at (\lx,\ly-4*\lsep) {Parametric DMP};
  \node[anchor=north west,font={\fontsize{5pt}{8}\selectfont}] at (\lx,\ly-5*\lsep) {PCA};
  \node[anchor=north west,font={\fontsize{5pt}{8}\selectfont}] at (\lx,\ly-6*\lsep) {RateInvariant};
  \node[anchor=north west] at (\lx+0.6,\ly+0.04) {\large \}};
  \node[anchor=north west] at (\lx+0.6,\ly-2*\lsep+0.04) {\large\}};
\end{tikzpicture}
  \caption{Test Distortion}
\end{subfigure}
  \caption{TimewarpVAE compared to beta-VAE, Parametric DMP,
  PCA, and Rate Invariant AE.
  Especially for lower-dimensional models,
  TimewarpVAE performs comparably to RateInvariant on training error and consistently outperforms RateInvariant in test error.}
  \label{fig:mainResult}
\end{figure*}

\begin{figure*}
\centering%
\begin{subfigure}[b]{0.3\textwidth}%
\centering%
\begin{tikzpicture}
  \useasboundingbox (-2.2,-1.6) rectangle (2.2,2);
  \node at (0,0) {\includegraphics[width=4cm]{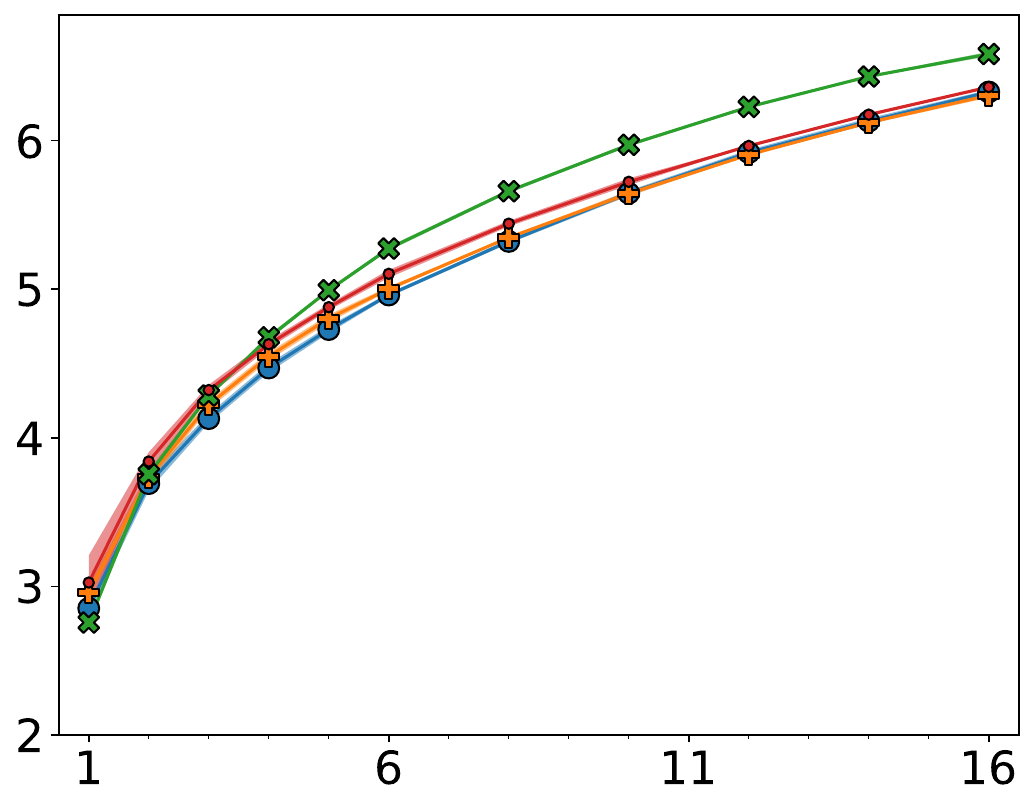}};
  \node at (0.15,-1.6) {\tiny Latent dimension};
  \node[anchor=center,rotate=90] at (-2.1,0) {\tiny Bits};
\end{tikzpicture}
  \caption{Rate}
\end{subfigure}
\begin{subfigure}[b]{0.3\textwidth}%
\centering%
\begin{tikzpicture}
  \useasboundingbox (-2.2,-1.6) rectangle (2.2,2);
  \node at (0,0) {\includegraphics[width=4cm]{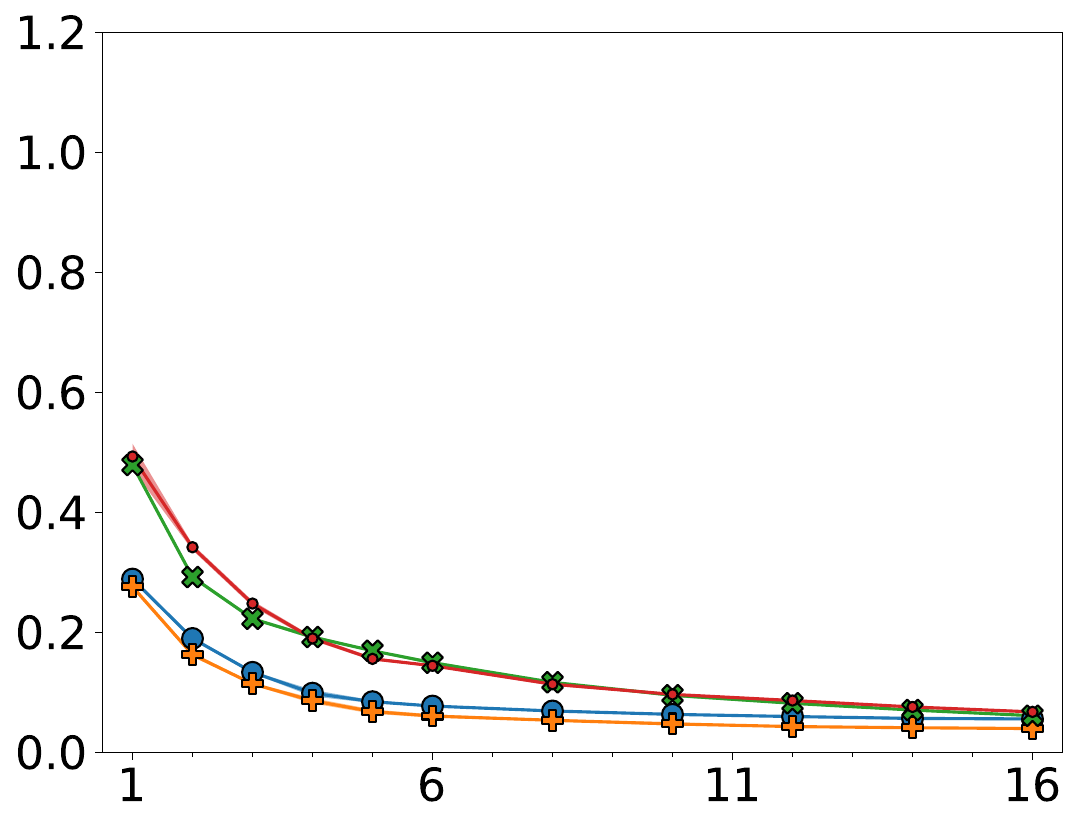}};
  \node at (0.15,-1.6) {\tiny Latent dimension};
  \node[anchor=center,rotate=90] at (-2.1,0) {\tiny A-RMSE};
\end{tikzpicture}
  \caption{Train Distortion}
\end{subfigure}
\begin{subfigure}[b]{0.3\textwidth}%
\centering%
\begin{tikzpicture}
  \useasboundingbox (-2.2,-1.6) rectangle (2.2,2);
  \node at (0,0) {\includegraphics[width=4cm]{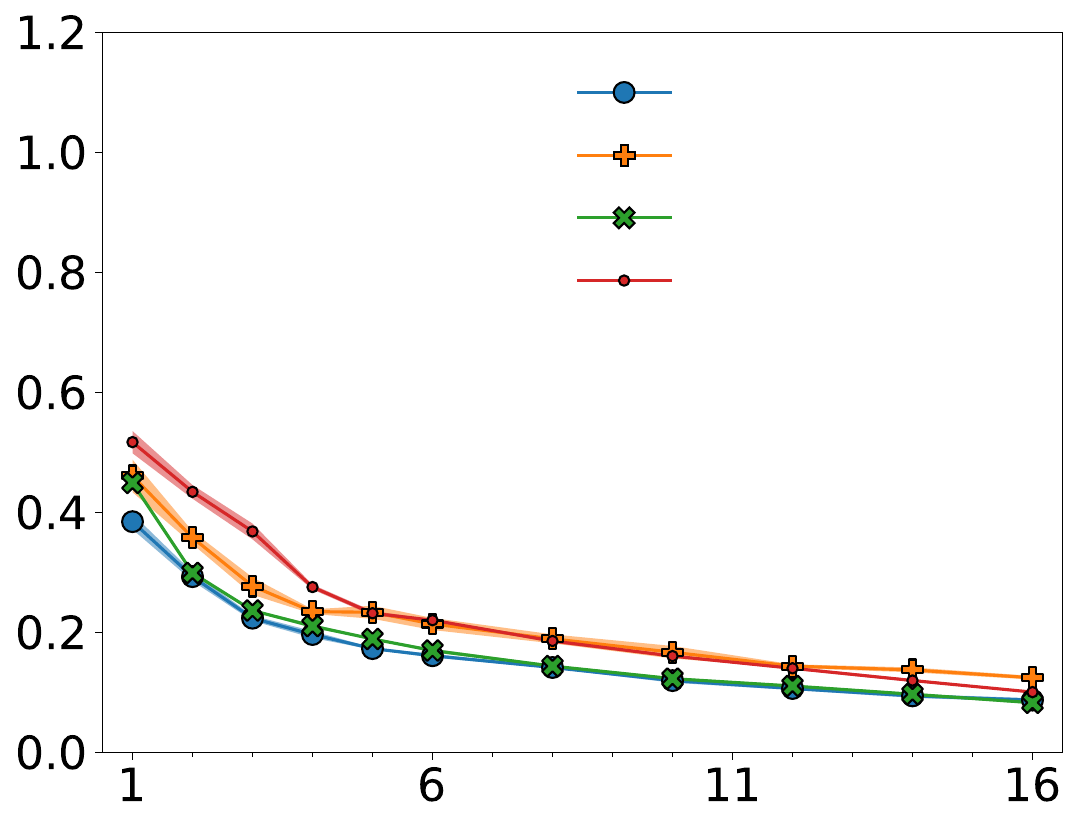}};
  \node at (0.15,-1.6) {\tiny Latent dimension};
  \node[anchor=center,rotate=90] at (-2.1,0) {\tiny A-RMSE};
  \def\lx{0.45}
  \def\ly{1.4}
  \node[anchor=north west,font={\fontsize{5pt}{8}\selectfont}] at (\lx,\ly) {TimewarpVAE};
  \node[anchor=north west,font={\fontsize{5pt}{8}\selectfont}] at (\lx,\ly-\lsep) {NoAugment};
  \node[anchor=north west,font={\fontsize{5pt}{8}\selectfont}] at (\lx,\ly-2*\lsep) {NoNonlinearity};
  \node[anchor=north west,font={\fontsize{5pt}{8}\selectfont}] at (\lx,\ly-3*\lsep) {NoTimewarp};
\end{tikzpicture}
  \caption{Test Distortion}
\end{subfigure}
  \caption{Ablation results show the data augmentation of timing noise is important for generalization performance, the nonlinear model gives a better fit to training data without losing generalization performance, and the time-warper is key to TimewarpVAE's good performance.}
  \label{fig:ablations}
\end{figure*}

Ablation studies were run to understand the importance of different architecture choices.
The first removes the data augmentation of additional trajectories with perturbed timings.
We call this ablation``NoAugment.''
The second removes the hidden layers in the neural network $\mathbf T(z)$.
Removing the hidden layers makes $\mathbf T(z)$ a linear function, meaning the function $f$ can only learn trajectories that
are a linear function of the latent variable $z$ (but $f$ is still nonlinear in the time argument $s$).
We call this ablation ``NoNonlinearity''.
The third removes the time-warper and replaces it with the identity function.
We call this ablation ``NoTimewarp.''
Results for these ablations are plotted in Fig.~\ref{fig:ablations}.
NoAugment confirms the importance of data augmentation for good generalization.
NoAugment can get a slightly better fit to the training data than TimewarpVAE, but performs poorly on test data.
NoNonlinearity has comparable performance on the test data to TimewarpVAE, 
showing the strict regularization imposed by its linearity condition is good for generalization.
However, NoNonlineary has such strong regularization that it cannot fit the training data as closely.
Additionally, the model rate for NoNonlinearity is higher. By constraining our model to be linear in the latent term, we cannot compress information as efficiently into the latent space.
Without the ability to time-align the data, 
NoTimewarping is not able to fit either the training or test data well.
The information rate 
of NoTimewarping is the same as that of TimewarpVAE, showing it does not compress spatial information as efficiently. 

\paragraph{Robot Execution.} We demonstrate the usefulness of our algorithm on a Kinova Gen3 robot arm.
Because of our timing regularization term, we can optimize various timing options during the replay of the trajectory, constraining
the trajectory to stay close to the demonstrated timings. This timing optimization allows the robot to consider its joint torque and speed limitations and choose a timing that it can execute the quickest under those constraints.
An example is shown in Fig.~\ref{fig:trajExecution}.
If the robot can only scale the timing uniformly, the optimized trajectory takes 1.8 times longer to execute. Our approach gives the robot the flexibility to slow down some parts of the trajectory and speed up other parts.

\section{Discussion}
TimewarpVAE is useful 
for simultaneously learning timing variations and latent factors of spatial variations.
Because it separately models timing variations,
TimewarpVAE concentrates the modeling capacity of its spatial latent variable on spatial factors of variation.
As discussed further in the Appendix, TimewarpVAE can be viewed
as a natural extension of continuous DTW.
TimewarpVAE uses a parametric model for the time-warping function.
A different approach
could be to use a non-parametric model like DTW to warp 
the reconstructed trajectory to the input trajectory and then backpropagate
the aligned error.
That alternate approach has a much higher computation cost 
because it requires computing DTW for each trajectory at each learning step and does not
re-use the time-warper from previous steps. We consider this approach briefly in the Appendix and leave it to future work to more fully understand the benefits of this alternate implementation and when it should be used or combined with the proposed TimewarpVAE.
This work measured the spatial error of reconstructed training and test trajectories, 
and showed
the TrajectoryVAE 
does a better job than beta-VAE at compressing spatial information into small latent spaces.
Future work will investigate also compressing the latent timing information.

\subsubsection*{Acknowledgments}
The authors would like to thank Sloan Nietert, Ziv Goldfeld, and Tapomayukh Bhattacharjee for valuable conversations and insights.

\subsubsection*{Reproducibility statement}
Our experiments are performed on a publicly available human gesture dataset of air-handwriting \cite{chen_6dmg_2012},
and on a dataset of quasistatic manipulation which we make publically available at
\url{https://github.com/travers-rhodes/TimewarpVAE}.
The PyTorch implementation of TimewarpVAE used in our experiments is also included at that url,
as well as the code to generate the figures for this paper.

\bibliography{ijcai24}
\bibliographystyle{named}

\appendix
\onecolumn
\section{Supplemental Materials}

\subsection{Derivation of TimewarpVAE from Dynamic Time Warping}
\label{sec:naturalExtensionCDTW}
Dynamic time warping (DTW) 
compensates for timing differences between two trajectories
by retiming the two trajectories so that 
they are spatially close to each other at matching timestamps.
In this section, we explicitly derive TimewarpVAE from continuous dynamic time warping, the formulation of dynamic time warping for continuous functions presented by \cite{1983SymmetricTimeWarping}.

\subsubsection{Continuous Dynamic Time Warping}
\label{sec:ContinuousDynamicTimeWarping}
We begin with a brief summary of continuous DTW.
Given two trajectories $x_0$ and $x_1$ (each a function from time in $[0,1]$ to some position in $\mathbb R^n$),
continuous DTW learns two time-warping functions,
$\rho_0$ and $\rho_1$, where each time-warping function is monotonic and bijective from $[0,1]$ to $[0,1]$.
The goal is to have
the time-warped trajectories be near each other at corresponding (warped) timesteps.
Mathematically, $\rho_0$ and $\rho_1$ are chosen to minimize the integral
\begin{equation}
  \int_0^1\left \| x_1\left(\rho_1(s)\right) - x_0\left(\rho_0(s)\right)\right\|^2
  \frac{
    \left(\rho_0\right)'\!\!(s) 
    +
    \left(\rho_1\right)'\!\!(s) 
  }{2}\,ds
\end{equation}
This integral is the distance between the trajectories at corresponding timesteps,
integrated over a symmetric weighting factor.

This algorithm has a degeneracy, in that many different $\rho_0$ and $\rho_1$ will lead to equivalent alignments $\rho_1\circ \rho_0^{-1}$
and will therefore have equal values of our cost function.
This becomes relevant when generating new trajectories from the interpolated model, 
as it requires choosing a timing
for the generated trajectory.

\subsubsection{Reformulation of Continuous DTW}
The optimization criterion of continuous DTW can be rewritten as follows:
Given the time-warping functions
$\rho_0$ and $\rho_1$ from above, 
define $\phi_0 = \rho_0^{-1}$ and $\phi_1=\rho_1^{-1}$.
Let the function $f : [0,1] \times [0,1] \rightarrow \mathbb R^n$
be defined as ${f(s,z) = (1-z) x_0(\rho_0(s)) + z x_1(\rho_1(t))}$.
That is, $f(s,z)$ is the unique function that is linear in its second parameter and which satisfies
the boundary conditions
${f(\phi_0(t),0) = x_0(t)}$ and ${f(\phi_1(t),1) = x_1(t)}$.
These boundary conditions associate $x_0$ with $z_0=0$ and $x_1$ with $z_1=1$.

    Our minimization objective for choosing $\rho_0$ and $\rho_1$ (or, equivalently, for choosing their inverses $\phi_0$ and $\phi_1$) can be written in terms of $f$ as
      \begin{equation}
        \frac 1 2 \int_0^1 \left \|\left.\frac {\partial f(s,z)}{\partial z}\right \rvert_{s=\phi_0(t),z=0} \right \|^2\,dt + 
        \frac 1 2 \int_0^1 \left \|\left.\frac {\partial f(s,z)}{\partial z}\right \rvert_{s=\phi_1(t),z=1}\right \|^2 dt.
      \end{equation}

The derivation goes as follows:
We define $\phi_0 = \rho^{-1}$ and $\phi_1=\rho^{-1}$,
and we define the function $f : [0,1] \times [0,1] \rightarrow \mathbb R^n$
to be the unique function that
is linear in its second parameter and which satisfies the boundary conditions 
    $f(\phi_0(t),0) = x_0(t)$ and 
    $f(\phi_1(t),1) = x_1(t)$.
Equivalently, it satisfies the boundary conditions
    $f(s,0) = x_0(\phi_0^{-1}(s))$ and 
    $f(s,1) = x_1(\phi_1^{-1}(s))$.

Substituting these definitions gives an optimization criterion of
\begin{equation}
  \int_0^1\left \| x_1\left(\phi_1^{-1}(s)\right) - x_0\left(\phi_0^{-1}(s)\right)\right\|^2
  \frac{
    \left(\phi_0^{-1}\right)'\!\!(s) 
    +
    \left(\phi_1^{-1}\right)'\!\!(s) 
  }{2}\,ds
\end{equation}
The boundary conditions of $f$ imply this is equal to

    \begin{equation}
      \frac 1 2 \int_0^1\left \| f(s,1) - f(s,0)\right\|
      d\phi_0^{-1}(s)
      +
      \frac 1 2 \int_0^1\left \| f(s,1) - f(s,0)\right\|
      d\phi_1^{-1}(s)
    \end{equation}
    And now, performing a change-of-variables $u = \phi_0^{-1}(s)$ and $v=\phi_1^{-1}(s)$ gives
    \begin{equation}
      \frac 1 2 \int_0^1\left \| f(\phi_0(u),1) - f(\phi_0(u),0)\right\|^2du
      +
      \frac 1 2 \int_0^1\left \| f(\phi_1(v),1) - f(\phi_1(v),0)\right\|^2dv
    \end{equation}
    Since $f$ is linear in its second coordinate, we can write this in terms of the partial derivatives of $f$
      \begin{equation}
        \frac 1 2 \int_0^1 \left \|\left.\frac {\partial f(s,z)}{\partial z}\right \rvert_{s=\phi_0(u),z=0} \right \|^2\,du + 
        \frac 1 2 \int_0^1 \left \|\left.\frac {\partial f(s,z)}{\partial z}\right \rvert_{s=\phi_1(v),z=1}\right \|^2 dv
      \end{equation}

\subsubsection{Simultaneous Time-Warping and Manifold Learning on Trajectories}
The relation to TimewarpVAE is as follows:

For each trajectory $x_i$, TimewarpVAE learns a low-dimensional latent representation $z_i\in \mathbb R^\ell$ associated with that trajectory. 
These $z_i$ are the natural generalizations of the reformulation of continuous DTW above, which had hard-coded latent values $z_0=0$ and $z_1=1$ for the two trajectories.

For each trajectory $x_i$, TimewarpVAE learns a time-warping function $\phi_i$ that transforms timesteps to new canonical timings.
These $\phi_i$ are the natural extension of the $\phi_0$ and $\phi_1$ from above.

TimewarpVAE learns a generative function $f$ which, given a canonical timestamp $s$ and a latent value $z$, returns the positon corresponding to the trajectory at that time.
This is an extension of the function $f$, with relaxations on the linearity constraint and the boundary conditions.
Instead of requiring $f$ to be linear in the $z$ argument, we parameterize $f$ with a neural network and regularize it to encourage $f$
to have small partial derivative with respect to the latent variable $z$.
This regularization is described in Section~\ref{sec:partialZRegularization}.
Instead of a boundary constraints requiring $f(\phi_i(t),z_i)$ to be equal to $x_i(t)$,
we instead add an optimization objective that $f(\phi_i(t),z_i)$ be close to $x_i(t)$,

\subsubsection{Regularization of the Decoder}
\label{sec:partialZRegularization}
Training the decoder using our optimization objective 
includes adding noise $\epsilon$ to the latent values $z$ before decoding.
This encourages the decoder to take on similar values for nearby values of $z$.
In particular, as described by \cite{kumar2020implicit},
this will add an implicit Jacobian squared regularization of the decoder over the $z$ directions. 
Penalizing these $\|\frac{\partial f(s,z)}{\partial z}\|^2$ terms is exactly what we want for our manifold-learning algorithm.
Additionally, we note that we do not add any noise to the temporal encoder when computing the reconstruction loss,
so our beta-VAE style architecture does not include any unwanted regularization of
$\|\frac{\partial f(s,z)}{\partial s}\|^2$.

\subsection[Degeneracy of Time Warping Using Notation of Weber]{Degeneracy of Time Warping Using Notation of \cite{Weber2019}}
\label{sec:DifferentFromWeber}
Using the notation of \cite{Weber2019}, the timing degeneracy noted in the main paper
also applies.
For any set of time-warping functions $T^{\theta_i}$, one for each of the $N_k$ different trajectories $u_i$ with class label $y_i$ out of $K$ possible class labels,
all of those time-warping functions can be composed with some additional fixed diffeomorphic warping $T^p$ without changing the Inverse Consistency Averaging Error.

That is, composing all time-warps with a time-warping function $T^p$ to generate $\tilde T^{\theta_i} = T^{\theta_i}\circ T^p$ will not affect the Inverse Consistency Averaging Error, since now the (perturbed) average warped trajectory for cluster $k$ $\tilde \mu_k$ is the warp of the previous average warped trajectory $\mu_k$
\begin{equation}
  \tilde \mu_k = \frac 1 {N_k} \sum u_i \circ T^{\theta_i}\circ T^p = \frac 1 {N_k} \left(\sum u_i \circ T^{\theta_i}\right) \circ T^p = \mu \circ T^p
\end{equation}.

The inverse $\tilde T^{-\theta_i}$ is simply $T^{-p} \circ T^{-\theta_i}$

The Inverse Consistency Averaging Error using those perturbed time-warps $\tilde T$ is the same as that computed using the original time-warps.
\begin{align}
  \mathcal L_{ICAE}(\tilde T) &= \sum_{k=1}^K \frac 1 {N_K} \sum_{i:y_i=k} \left \| \tilde \mu_k \circ \tilde T^{\theta_i} - u_i \right \|_{\ell_2}^2 \\
                    &= \sum_{k=1}^K \frac 1 {N_K} \sum_{i:y_i=k} \left \| \mu_k \circ T^p \circ T^{-p} \circ T^{-\theta_i} - u_i \right \|_{\ell_2}^2\\
                    &= \sum_{k=1}^K \frac 1 {N_K} \sum_{i:y_i=k} \left \| \mu_k \circ T^{-\theta_i} - u_i \right \|_{\ell_2}^2 \\
                    &= \mathcal L_{ICAE}(T)
\end{align}

This shows that there is a degeneracy over the choice of warping of the warped trajectories. 
Warped trajectories can all be additionally warped by another time-warping function without changing the Inverse Consistency Averaging Error.

\subsection{Plot of time-warping basis function}
\label{sec:timeAlignmentBasisFunctionsPlot}
\begin{figure}
  \center
  \begin{tikzpicture}
    \def\isep{1.12}
    \node at (0,0) {\includegraphics[width=8cm]{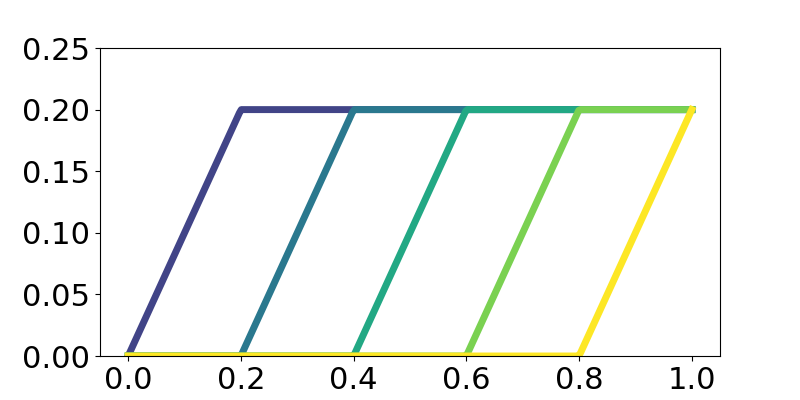}};
    \node[rotate=90] at (-4,0) {$\psi_j(t)$};
    \foreach \i in {1,...,5}{
      \node[rotate=65] at ($(-2.25 + \i * \isep - \isep,0)$) {$i=\i$};
    }
    \node at (0,-2) {$t$};
  \end{tikzpicture}
  \caption{Time alignment basis functions for $K=5$, for each $i$ from $1$ to $5$}
  \label{fig:timeAlignmentBasisFunctions}
\end{figure}

A plot of the time-warping basis functions $\psi$ is shown in Fig.~\ref{fig:timeAlignmentBasisFunctions}.

\subsection{Example training trajectories of letter ``A''}
\label{sec:exampleATrajs}
\begin{figure}%
\centering%
\begin{tikzpicture}
  \def\spac{2.2}
  \foreach \i in {0,...,4} {
    \node[draw,thick,inner sep=0, outer sep=0] at ($(\i*\spac,0)$){\includegraphics[width=\spac cm]{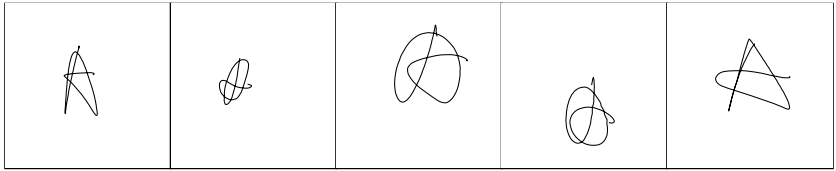}};
  }
\end{tikzpicture}
  \caption{Five example trajectories of handwritten ``A''}
  \label{fig:exampleAs}
\end{figure}

Example training trajectories of handwritten letter ``A'''s are shown in Fig.~\ref{fig:exampleAs}.

\subsection{Time-Warper Effect on Trained Model}
\begin{figure}%
\centering%
\begin{subfigure}[b]{0.25\textwidth}%
\centering%
\begin{tikzpicture}
  \node at (0,0) (bigpic) {\includegraphics[width=\textwidth, trim=0 70 0 0, clip]{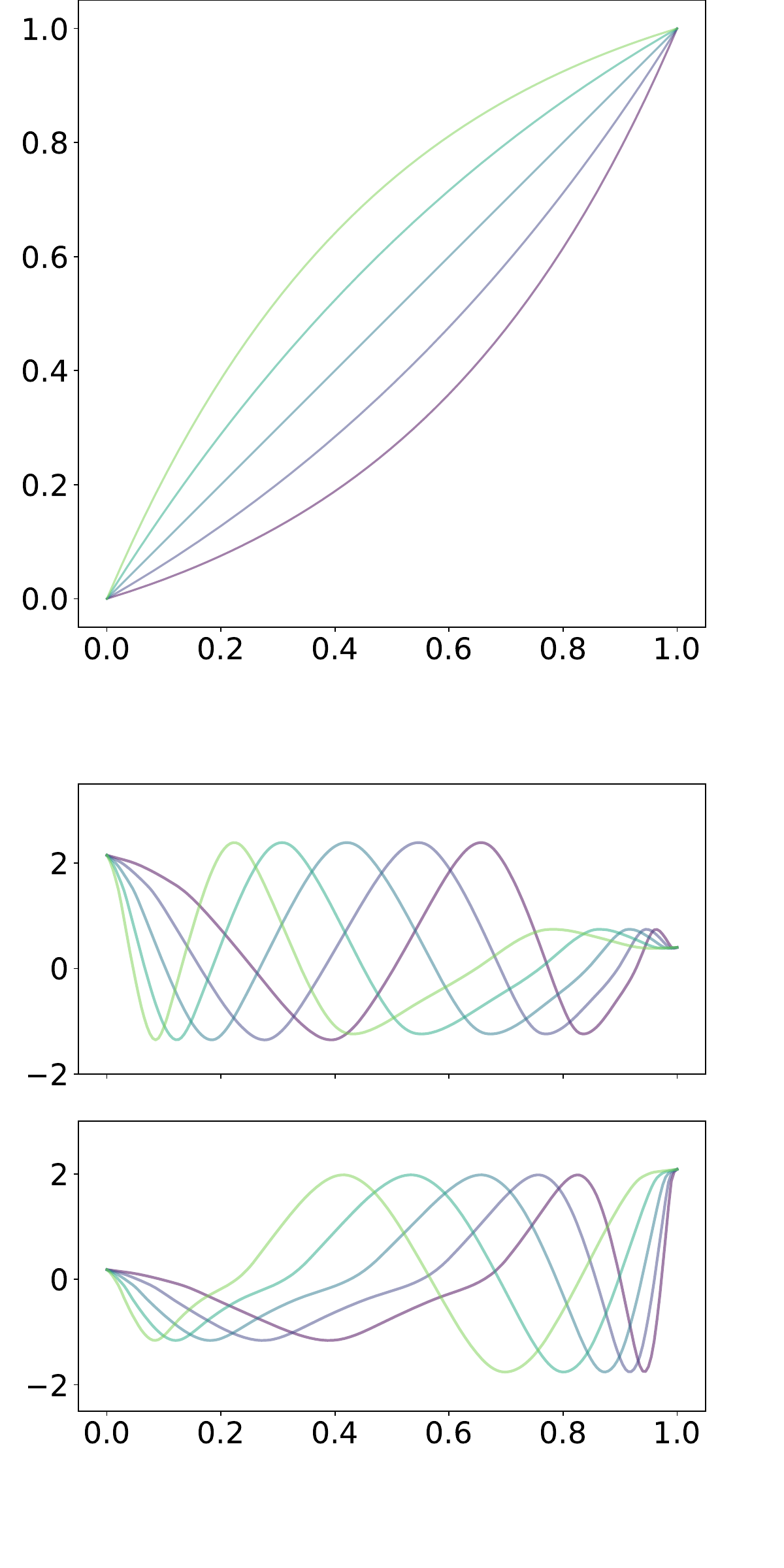}};
    \node[fill=white,minimum width=1cm,minimum height=0.1cm,inner sep=0, outer sep=0] 
      at ($(bigpic.center) + (\coff,\figxlaba)$) {\tiny Time t};
    \node[fill=white,minimum width=1cm,minimum height=0.1cm,anchor=center,rotate=90,inner sep=0, outer sep=0] 
      at ($(bigpic.west) + (0,\figylaba)$) {\tiny Retimed $\phi(t)$};
    \node[fill=white,minimum width=1cm,minimum height=0.1cm,inner sep=0, outer sep=0] 
      at ($(bigpic.south)+(\coff,\figxlabb)$) {\tiny Time t};
    \node[fill=white,minimum width=1.3cm,minimum height=0.1cm,anchor=center,rotate=90,inner sep=0, outer sep=0] 
      at ($(bigpic.west) + (0,\figylabb)$) {\tiny Y Position};
    \node[fill=white,minimum width=1.3cm,minimum height=0.1cm,anchor=center,rotate=90,inner sep=0, outer sep=0] 
      at ($(bigpic.west) + (0,\figylabc)$) {\tiny X Position};
\end{tikzpicture}%
\end{subfigure}
\caption{Decoding the same spatial latent variable using different time-warping parameters will give the same spatial
  trajectory but with different timings (fast or slow at different times). We plot the time-warping functions for five different timing latents
  and the generated trajectories for a single spatial latent value by showing the generated positions as a function of time.}%
\label{fig:timewarpEffect}%
\end{figure}

In Fig.~\ref{fig:timewarpEffect} we show how varying the time-warper parameters affects the timing of the generated trajectory
for one of the TimewarpVAE letter models.
Here, we choose five different sets of time-warper parameters, and 
plot the resulting time-warper function $\phi(t)$.
We then decode the same spatial trajectory using those five latent parameters.
The resulting trajectories spatially would look all the same, and we plot the X and Y locations of the generated trajectories as a function of time.
The associated timings of the trajectories changes due to the time-warper function.

\subsection{Additional Interpolations for Models}
\label{sec:addlInterpolations}
\begin{figure}%
\centering%
\begin{subfigure}[b]{0.20\textwidth}%
\centering%
\begin{tikzpicture}
  \node at (0,0) (bigpic) {\includegraphics[width=\textwidth, trim=20 70 0 0, clip]{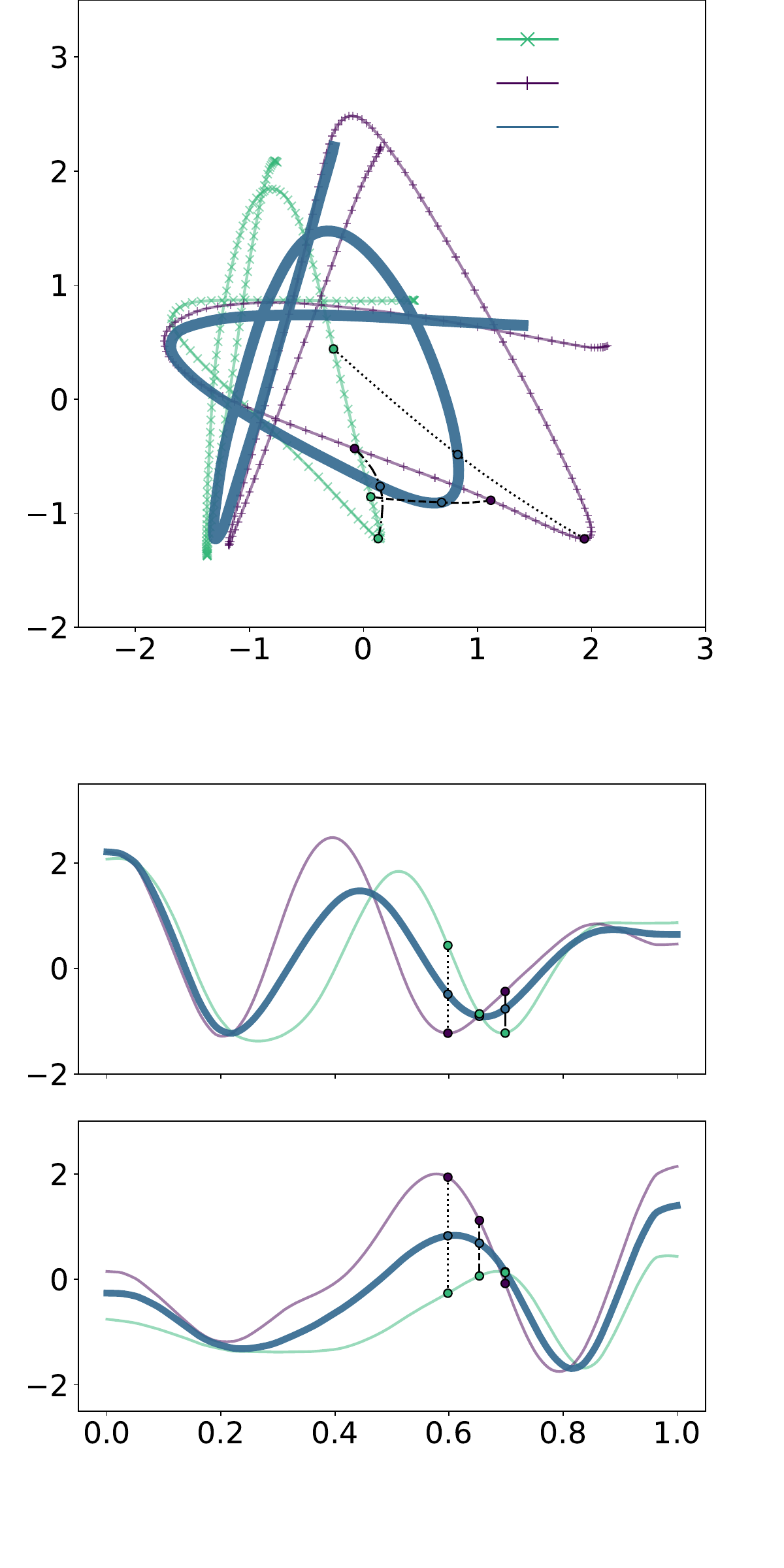}};
    \node[fill=white,minimum width=1cm,minimum height=0.1cm,inner sep=0, outer sep=0] 
      at ($(bigpic.center) + (\coff,\figxlaba)$) {\tiny X Position};
    \node[fill=white,minimum width=1cm,minimum height=0.1cm,anchor=center,rotate=90,inner sep=0, outer sep=0] 
      at ($(bigpic.west) + (0,\figylaba)$) {\tiny Y Position};
    \node[fill=white,minimum width=1cm,minimum height=0.1cm,inner sep=0, outer sep=0] 
      at ($(bigpic.south)+(\coff,\figxlabb)$) {\tiny Canonical Time};
    \node[fill=white,minimum width=1.3cm,minimum height=0.1cm,anchor=center,rotate=90,inner sep=0, outer sep=0] 
      at ($(bigpic.west) + (0,\figylabb)$) {\tiny Y Position};
    \node[fill=white,minimum width=1.3cm,minimum height=0.1cm,anchor=center,rotate=90,inner sep=0, outer sep=0] 
      at ($(bigpic.west) + (0,\figylabc)$) {\tiny X Position};
  \node[anchor=north west,font={\fontsize{5pt}{8}\selectfont}] at (\interplx,\interply) {\trajlaba};
  \node[anchor=north west,font={\fontsize{5pt}{8}\selectfont}] at (\interplx,\interply-\legendysep) {\trajlabb};
  \node[anchor=north west,font={\fontsize{5pt}{8}\selectfont}] at (\interplx,\interply-2*\legendysep) {\trajlabc};
\end{tikzpicture}%
\caption{NoTimewarp}%
\label{fig:noTimewarpAverage}%
\end{subfigure}%
\hfill%
\begin{subfigure}[b]{0.20\textwidth}%
\centering%
\begin{tikzpicture}
  \node at (0,0) (bigpic) {\includegraphics[width=\textwidth, trim=20 70 0 0, clip]{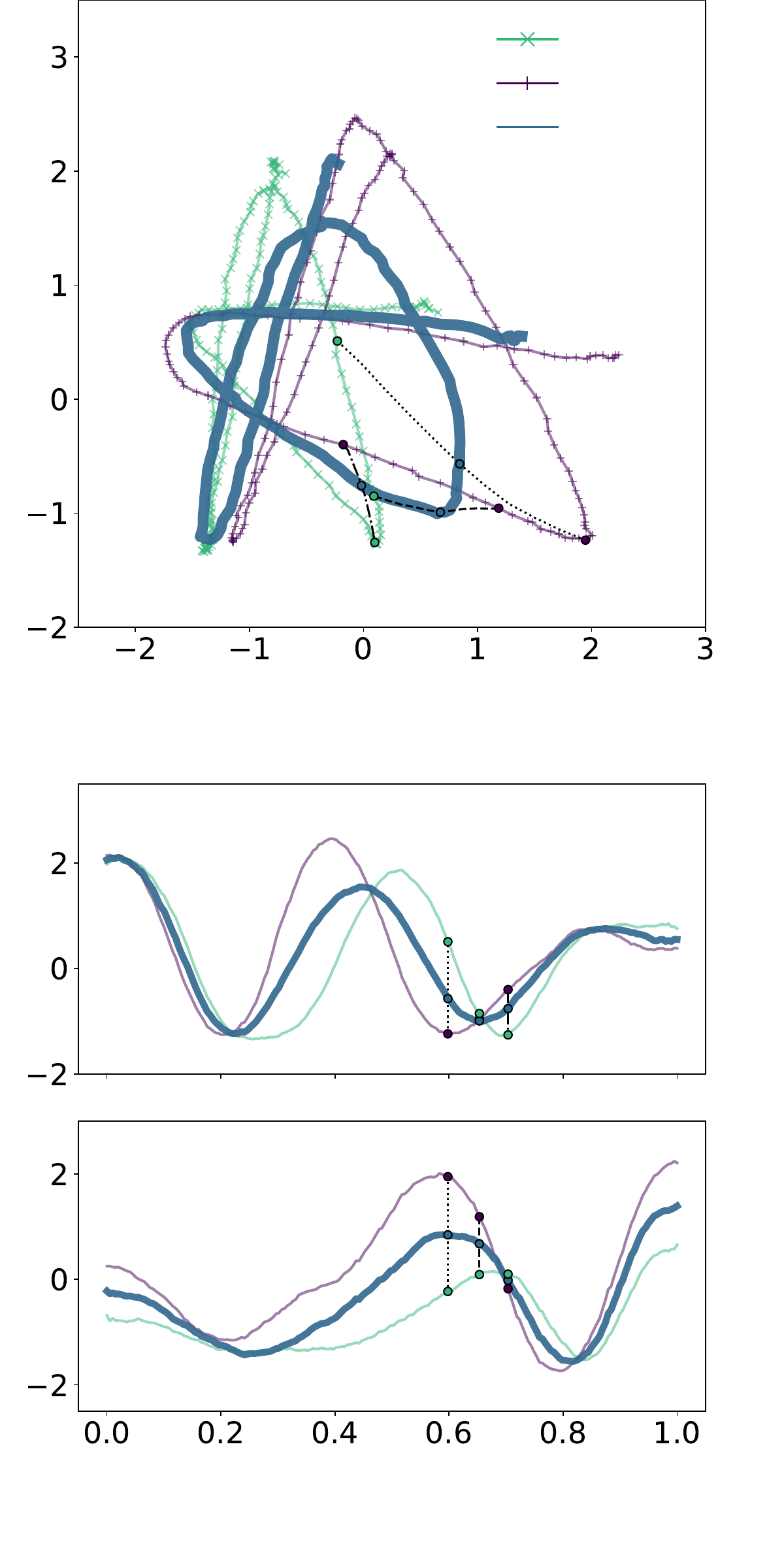}};
    \node[fill=white,minimum width=1cm,minimum height=0.1cm,inner sep=0, outer sep=0] 
      at ($(bigpic.center) + (\coff,\figxlaba)$) {\tiny X Position};
    \node[fill=white,minimum width=1cm,minimum height=0.1cm,anchor=center,rotate=90,inner sep=0, outer sep=0] 
      at ($(bigpic.west) + (0,\figylaba)$) {\tiny Y Position};
    \node[fill=white,minimum width=1cm,minimum height=0.1cm,inner sep=0, outer sep=0] 
      at ($(bigpic.south)+(\coff,\figxlabb)$) {\tiny Canonical Time};
    \node[fill=white,minimum width=1.3cm,minimum height=0.1cm,anchor=center,rotate=90,inner sep=0, outer sep=0] 
      at ($(bigpic.west) + (0,\figylabb)$) {\tiny Y Position};
    \node[fill=white,minimum width=1.3cm,minimum height=0.1cm,anchor=center,rotate=90,inner sep=0, outer sep=0] 
      at ($(bigpic.west) + (0,\figylabc)$) {\tiny X Position};
  \node[anchor=north west,font={\fontsize{5pt}{8}\selectfont}] at (\interplx,\interply) {\trajlaba};
  \node[anchor=north west,font={\fontsize{5pt}{8}\selectfont}] at (\interplx,\interply-\legendysep) {\trajlabb};
  \node[anchor=north west,font={\fontsize{5pt}{8}\selectfont}] at (\interplx,\interply-2*\legendysep) {\trajlabc};
\end{tikzpicture}%
\caption{beta-VAE}%
\label{fig:convVAEAverage}%
\end{subfigure}%
\hfill%
\begin{subfigure}[b]{0.20\textwidth}%
\centering%
\begin{tikzpicture}
  \node at (0,0) (bigpic) {\includegraphics[width=\textwidth, trim=20 70 0 0, clip]{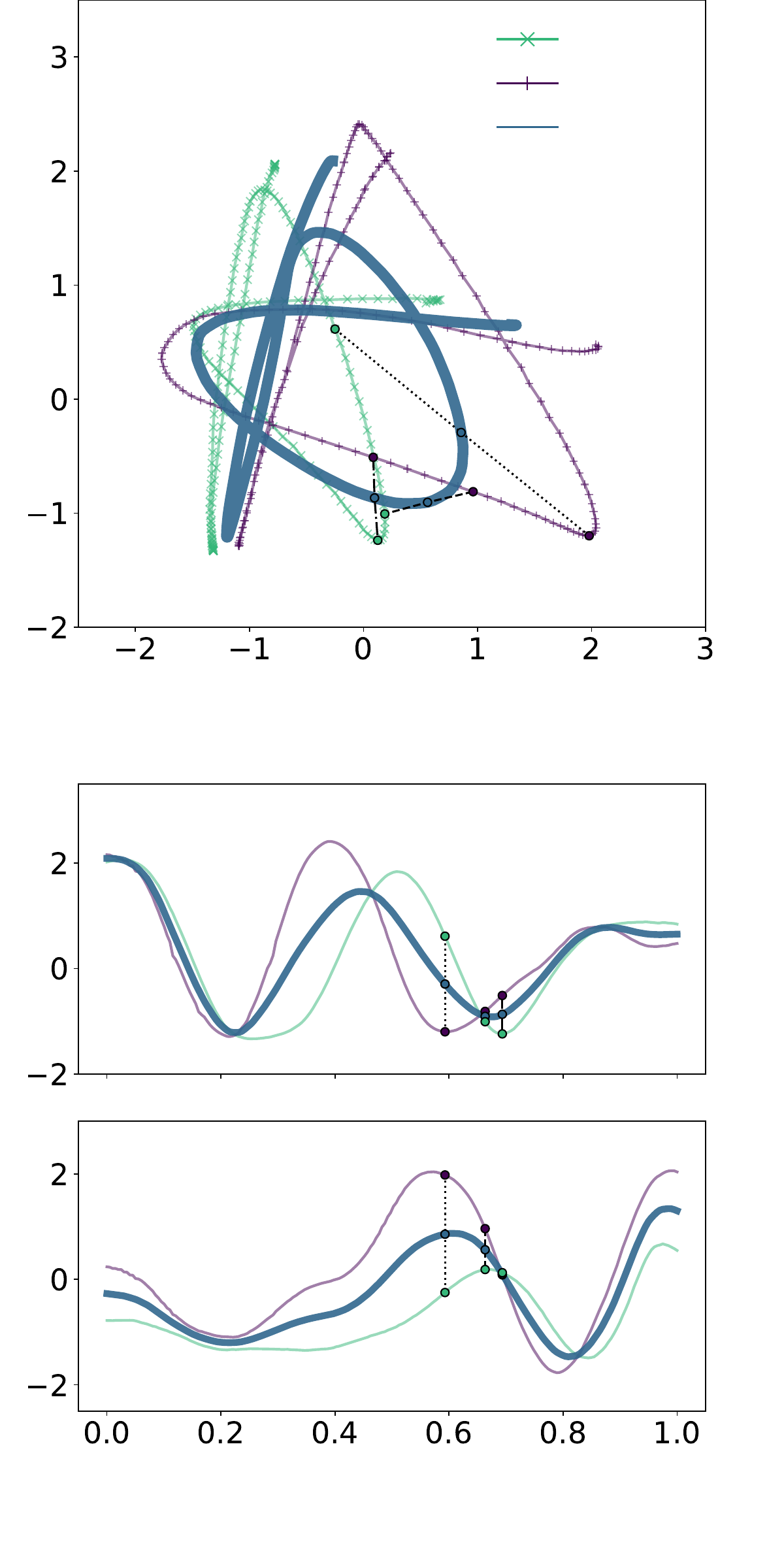}};
    \node[fill=white,minimum width=1cm,minimum height=0.1cm,inner sep=0, outer sep=0] 
      at ($(bigpic.center) + (\coff,\figxlaba)$) {\tiny X Position};
    \node[fill=white,minimum width=1cm,minimum height=0.1cm,anchor=center,rotate=90,inner sep=0, outer sep=0] 
      at ($(bigpic.west) + (0,\figylaba)$) {\tiny Y Position};
    \node[fill=white,minimum width=1cm,minimum height=0.1cm,inner sep=0, outer sep=0] 
      at ($(bigpic.south)+(\coff,\figxlabb)$) {\tiny Canonical Time};
    \node[fill=white,minimum width=1.3cm,minimum height=0.1cm,anchor=center,rotate=90,inner sep=0, outer sep=0] 
      at ($(bigpic.west) + (0,\figylabb)$) {\tiny Y Position};
    \node[fill=white,minimum width=1.3cm,minimum height=0.1cm,anchor=center,rotate=90,inner sep=0, outer sep=0] 
      at ($(bigpic.west) + (0,\figylabc)$) {\tiny X Position};
  \def\lx{0.6}
  \def\ly{3.45}
  \node[anchor=north west,font={\fontsize{5pt}{8}\selectfont}] at (\interplx,\interply) {\trajlaba};
  \node[anchor=north west,font={\fontsize{5pt}{8}\selectfont}] at (\interplx,\interply-\legendysep) {\trajlabb};
  \node[anchor=north west,font={\fontsize{5pt}{8}\selectfont}] at (\interplx,\interply-2*\legendysep) {\trajlabc};
\end{tikzpicture}%
\caption{DMP}%
\label{fig:dmpAverage}%
\end{subfigure}%
\hfill%
\begin{subfigure}[b]{0.20\textwidth}%
\centering%
\begin{tikzpicture}
  \node at (0,0) (bigpic) {\includegraphics[width=\textwidth, trim=20 70 0 0, clip]{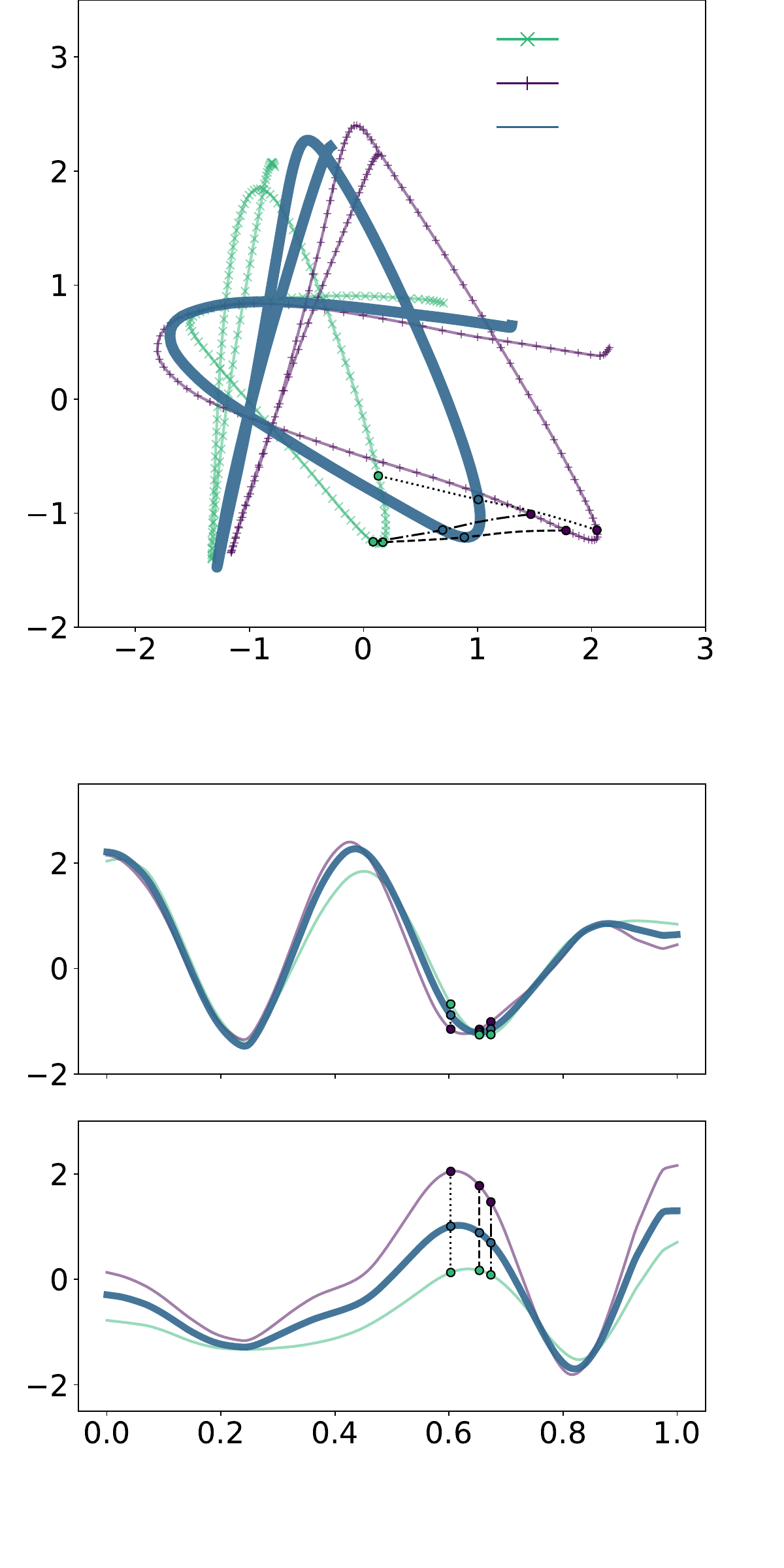}};
    \node[fill=white,minimum width=1cm,minimum height=0.1cm,inner sep=0, outer sep=0] 
      at ($(bigpic.center) + (\coff,\figxlaba)$) {\tiny X Position};
    \node[fill=white,minimum width=1cm,minimum height=0.1cm,anchor=center,rotate=90,inner sep=0, outer sep=0] 
      at ($(bigpic.west) + (0,\figylaba)$) {\tiny Y Position};
    \node[fill=white,minimum width=1cm,minimum height=0.1cm,inner sep=0, outer sep=0] 
      at ($(bigpic.south)+(\coff,\figxlabb)$) {\tiny Canonical Time};
    \node[fill=white,minimum width=1.3cm,minimum height=0.1cm,anchor=center,rotate=90,inner sep=0, outer sep=0] 
      at ($(bigpic.west) + (0,\figylabb)$) {\tiny Y Position};
    \node[fill=white,minimum width=1.3cm,minimum height=0.1cm,anchor=center,rotate=90,inner sep=0, outer sep=0] 
      at ($(bigpic.west) + (0,\figylabc)$) {\tiny X Position};
  \def\lx{0.6}
  \def\ly{3.45}
  \node[anchor=north west,font={\fontsize{5pt}{8}\selectfont}] at (\interplx,\interply) {\trajlaba};
  \node[anchor=north west,font={\fontsize{5pt}{8}\selectfont}] at (\interplx,\interply-\legendysep) {\trajlabb};
  \node[anchor=north west,font={\fontsize{5pt}{8}\selectfont}] at (\interplx,\interply-2*\legendysep) {\trajlabc};
\end{tikzpicture}%
\caption{TimewarpVAE-DTW}%
\label{fig:timewarpVAEDTWAverage}%
\end{subfigure}%
\caption{Additional interpolation results}%
\end{figure}

Here, we present additional interpolation results, all on 16 latent dimensions and $\beta=0.001$.
We note that the convolutional encoder/decoder architecture in beta-VAE in Fig.~\ref{fig:convVAEAverage} 
does not appear to have as strong an implicit bias toward smooth trajectories as the TimewarpVAE architecture.
This makes sense because the TimewarpVAE architecture decomposes the generative function into a component $g(s)$ which
computes poses as a function of time, likely inducing an inductive bias toward smoother trajectories as a function of time.

The interpolation in Fig.~\ref{fig:noTimewarpAverage} shows that in the ablation of TimewarpVAE without the timing module
the interpolation does not preserve the style of the ``A''.
Likewise, the DMP interpolation in Fig.~\ref{fig:dmpAverage} 
does not preserve the style of the ``A''.

\subsection{Equivalence of Regularizing $\phi$ or $\phi^{-1}$}
\label{sec:equivInverse}
We note that our regularization is the same, regardless if we regularize $\phi$ or $\phi^{-1}$.
We show this by applying the substitution $s = \phi(t)$, $ds = \phi'(t)dt$.
That substitution gives:
$\int_0^1 \left(1 -\frac{1}{\phi'(\phi^{-1}(s))}\right) \log\left( \phi'(\phi^{-1}(s))\right)ds$.

We use the identity
$(\phi^{-1})'(s) = \frac{1}{\phi'(\phi^{-1}(s))}$
to simplify to

\begin{equation}
  \int_0^1 \left(1 -(\phi^{-1})'(s)\right) \log\left( \frac 1 {(\phi^{-1})'(s)}\right)ds
  = \int_0^1 \left((\phi^{-1})'(s) - 1\right) \log\left( (\phi^{-1})'(s)\right)ds
\end{equation}

This is exactly our regularization applied to the function $\phi^{-1}$.
Thus, we note that our regularization is symmetric. Our regularization cost is the same whether it is applied to $\phi$ (the function from trajectory time to canonical time) or 
applied to $\phi^{-1}$ (the function from canonical time to trajectory time).

It didn't have to be that way.
For example, if our cost function were of the form $\int_0^1 \log^2(\phi'(t)) dt$, the substitution above would give a cost
$\int_0^1 \frac{1}{\phi'(\phi^{-1}(s))}\log^2(\phi'(\phi^{-1}(s))) ds$
which simplifies to
$\int_0^1 \frac{1}{\phi'(\phi^{-1}(s))}\log^2((\phi^{-1})'(s)) ds$
which equals
$\int_0^1 (\phi^{-1})'(s)\log^2((\phi^{-1})'(s)) ds$
which is different from applying the regularization procedure to $\phi^{-1}$ which would have given a regularization term of:
$\int_0^1 \log^2((\phi^{-1})'(s)) ds$

\subsection{Inspiration for Regularization Function}
Our regularization function was inspired by Unbalanced Optimal Control.
If we assign a uniform measure $\mathcal U$ to $[0,1]$,
we note that our regularization cost is exactly
the symmetric KL Divergenge between $\mathcal U$ and the pushforward ${\phi_i}_* \mathcal U$.
For each $\phi_i$, the pushforward ${\phi_i}_*\mathcal U$ has a probability density function $1/(\phi_i'(\phi_i^{-1}(s)))$,
and the symmetric KL divergence cost
${D_{\mathrm{KL}}(\mathcal U \vert {\phi_i}_*\mathcal U) + 
D_{\mathrm{KL}}({\phi_i}_*\mathcal U \vert \mathcal U )}$ gives the loss given above.

We work through the explicit mathematics below.

\subsubsection{Pushforward of Probability Density Function}
If $F_0$ is some cumulative distribution function, and $F_1$ is the CDF generated by the pushforward of a function $\phi$ then we have the simple identity $F_1(\phi(t)) = F_0(t)$.
Taking the derivative of both sides with respect to $t$ gives $F_1'(\phi(t))\phi'(t) = F_0'(t)$.
The substitutions $s = \phi(t)$ and $\phi^{-1}(s) = t$ give
${F_1'(s) =\frac 1 {\phi'(\phi^{-1}(s))} F_0'(\phi^{-1}(s))}$.
Since the PDF is the derivative of the CDF, then writing $f_0$ and $f_1$ as the corresponding PDFs of $F_0$ and $F_1$, we see 
\begin{equation}
  f_1(s) =\frac 1 {\phi'(\phi^{-1}(s))} f_0(\phi^{-1}(s))
\end{equation}
In the simple case where $f_0 = \mathcal U$, the uniform distribution over $[0,1]$, then $f_0(t) = 1$, so we have our pushforward
\begin{equation}
  \left(\phi_* \mathcal U\right)(s) = \frac 1 {\phi'(\phi^{-1}(s))} 
\end{equation}

\subsubsection{Symmetric KL Divergence Calculation}
The KL divergence $D_{\mathrm{KL}}(\mathcal U \vert {\phi}_*\mathcal U)$ is 
$\int_0^1 \log\left( \phi'(\phi^{-1}(s))\right)ds$.
Substituting $\phi(t) = s$ and the corresponding $\phi'(t)dt = ds$ gives 
${\int_0^1 \phi'(t)\log\left( \phi'(t)\right)dt}$.
Likewise, 
the KL divergence $D_{\mathrm{KL}}({\phi}_*\mathcal U \vert \mathcal U)$ is
$\int_0^1 \frac 1 {\phi'(\phi^{-1}(s))}\log\left(\frac 1 {\phi'(\phi^{-1}(s))}\right)ds$, 
which simplifies with the same substitutions to 
$-\int_0^1 \log\left( \phi'(t)\right)dt$.

The symmetric KL divergence cost is thus
\begin{equation}
D_{\mathrm{KL}}(\mathcal U \vert {\phi}_*\mathcal U) + 
D_{\mathrm{KL}}({\phi}_*\mathcal U\vert\mathcal U ) =
  \int_0^1 \left(\phi'(t)-1\right) \log\left( \phi'(t)\right)dt
\end{equation}

\subsection{Initialization of Neural Network for $f(s,z)$}
\label{sec:customInit}
\begin{figure}
  \centering%
  \begin{subfigure}[b]{0.48\textwidth}
    \begin{tikzpicture}
      \node at (0,0) {\includegraphics[width=\linewidth]{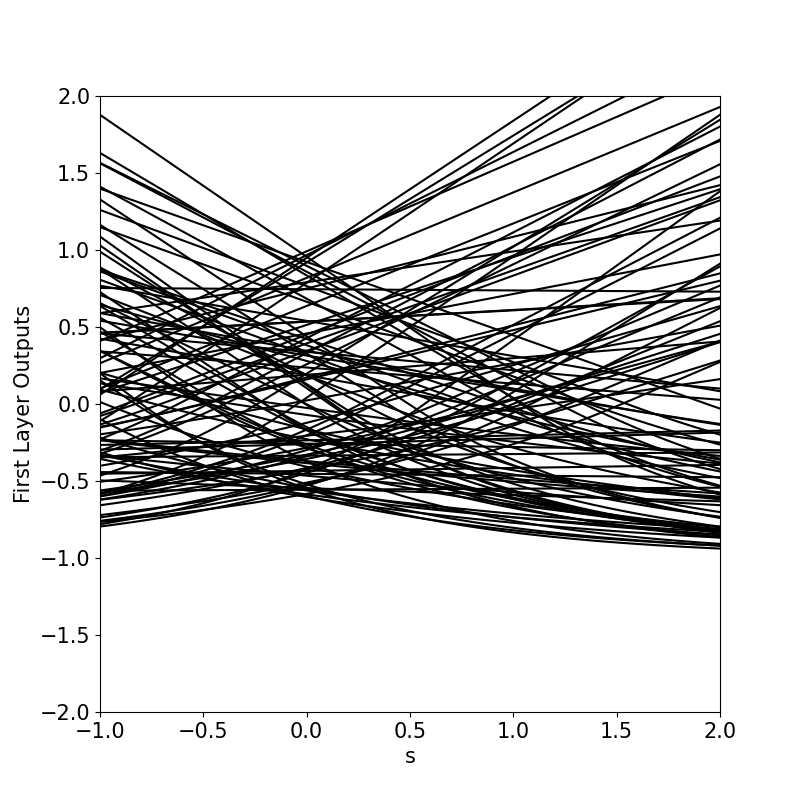}};
    \end{tikzpicture}%
    \caption{Default PyTorch Initialization}
  \end{subfigure}
  \hfill
  \begin{subfigure}[b]{0.48\textwidth}
    \begin{tikzpicture}
      \node at (0,0) {\includegraphics[width=\linewidth]{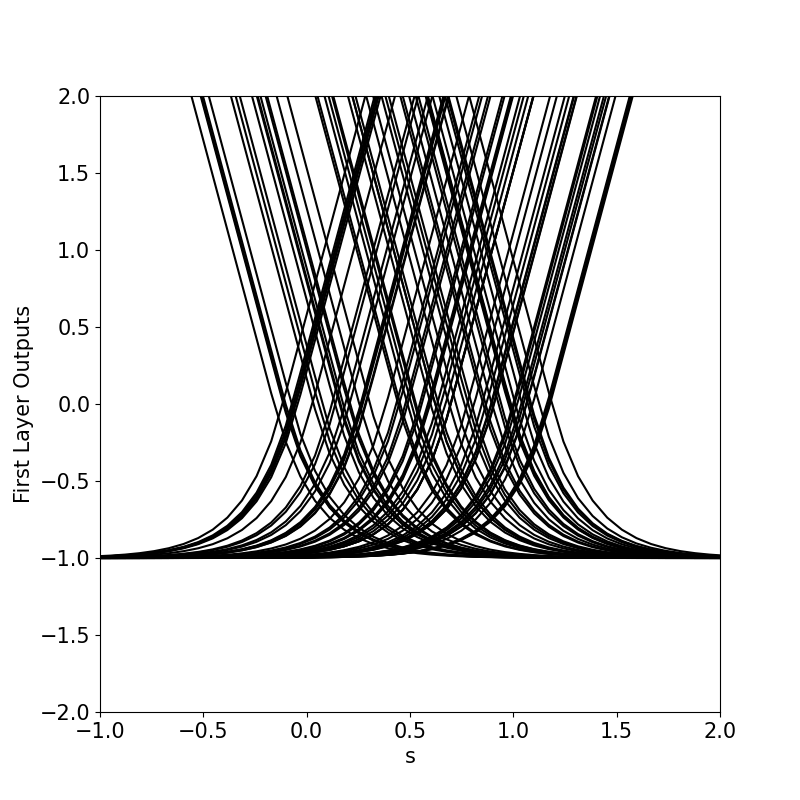}};
    \end{tikzpicture}%
    \caption{Custom Initialization}
  \end{subfigure}
  \begin{caption}
    {
      Outputs of first layer of the neural network $g(s)$. 
    }
  \end{caption}
\end{figure}

As mentioned above, the neural network $f(s,z)$ is split into $\mathbf T(z)$ and $g(s)$. 
Since $g(s)$ takes in a canonical time $s\in [0,1]$ which we want to have roughly uniform modeling capacity over 
the full range of $s$ from $0$ and $1$,
we initialize the first layer of the neural network's weights, $W$ (a matrix with one column), and bias $b$ (a vector) for $g(s)$ in the following way.

We initialize the values in $W$
to be randomly, independently $-G$ or $G$ with equal probability,
where $G$ is a hyperparameter.

We then choose values of $b$ so that,
for each (output) row $j$, which we denote $W_j$ and $b_j$
the y-intercepts of the function $y = g_j(s) = W_j s + b_j$
are each an independent uniformly random value in $[0-\eta, 1+\eta]$

Visually, the effect of this initialization can be seen by plotting 
the first layer's transformation $\mathrm{ELU}(W s + b)$
using PyTorch's default initialization 
and using our proposed initialization,
where ELU is the Exponential Linear Unit introduced by \cite{ClevertUH15}.
PyTorch's default implementation randomly initializes the output functions to be distributed symmetrically around $s=0$. 
Additionally, much of the modeling capacity is assigned to variations outside the domain $[0,1]$.
Since we know that the input timestamp will be $s\in[0,1]$,
our initialization focuses the modeling capacity near $[0,1]$ and is symmetric around the middle of that range $s=0.5$.

\subsection{TimewarpVAE Timing Noise Data Augmentation}
\label{sec:noiseAugmentation}
\begin{figure}
  \centering
  \begin{tikzpicture}
    \node at (0,0) (plot) {\includegraphics[width=6cm]{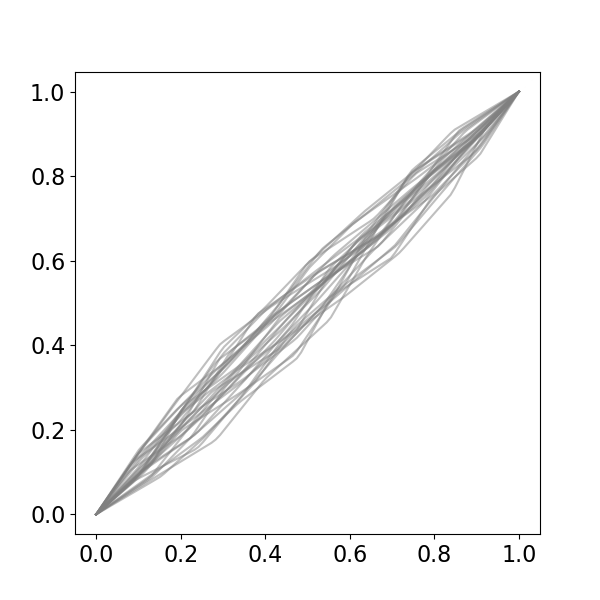}};
    \node at (plot.south) {Timestamp};
    \node[rotate=90] at (plot.west) {Timestamp with added noise};
  \end{tikzpicture}
  \caption{Example (random) functions used to add timing noise during data augmentation}
  \label{fig:noiseAugmentation}
\end{figure}

For data augmentation of timing noise, we create perturbed timesteps to sample the training trajectories as follows.
First, we construct two random vectors $\nu_{\text{in}}$ and $\nu_{\text{out}}$ of uniform random numbers between $0$ and $1$,
each vector of length $10$.
We then square each of the elements in those vectors and multiply by a noise hyperparameter $\eta$
and take the cumulative sum to give perturbation vectors for input and output timings.
We add each of those vectors to a vector with ten elements with linear spacing, $[0,1/9,2/9,3/9,\ldots,1]$.
We then normalize those perturbed vectors so that the last elements are again $1$.
The resulting vectors now give nicely symmetric, monotonic x and y coordinates of knots for a stepwise-linear perturbation vector which we can subsample at arbitrary
timesteps to give desired noise-added output timesteps.
We choose $\eta=0.1$ in our experiments, giving noise functions plotted in Fig.~\ref{fig:noiseAugmentation}.

When using this data augmentation, each time we pass a training trajectory into our model, instead of sampling
the training trajectory at $T$ uniformly-distributed timesteps between $0$ and $1$ to construct our $x \in \mathbb R^{T\times n}$,
we first perturb all those timesteps by passing them through a randomly generated noise functions.
This means that each $x$ we pass into our learning algorithm has a slightly different timing than the training data,
allowing us to perform data augmentation on the timings of the training data.

\subsection{Alternative TimewarpVAE Approach Directly Using Dynamic Time Warping}
\label{sec:TimewarpVAEDTW}
\begin{figure}
\centering%
\begin{subfigure}[b]{0.3\textwidth}%
\centering%
\begin{tikzpicture}
  \useasboundingbox (-2.2,-2) rectangle (2.2,2);
  \node at (0,0) {\includegraphics[width=4cm]{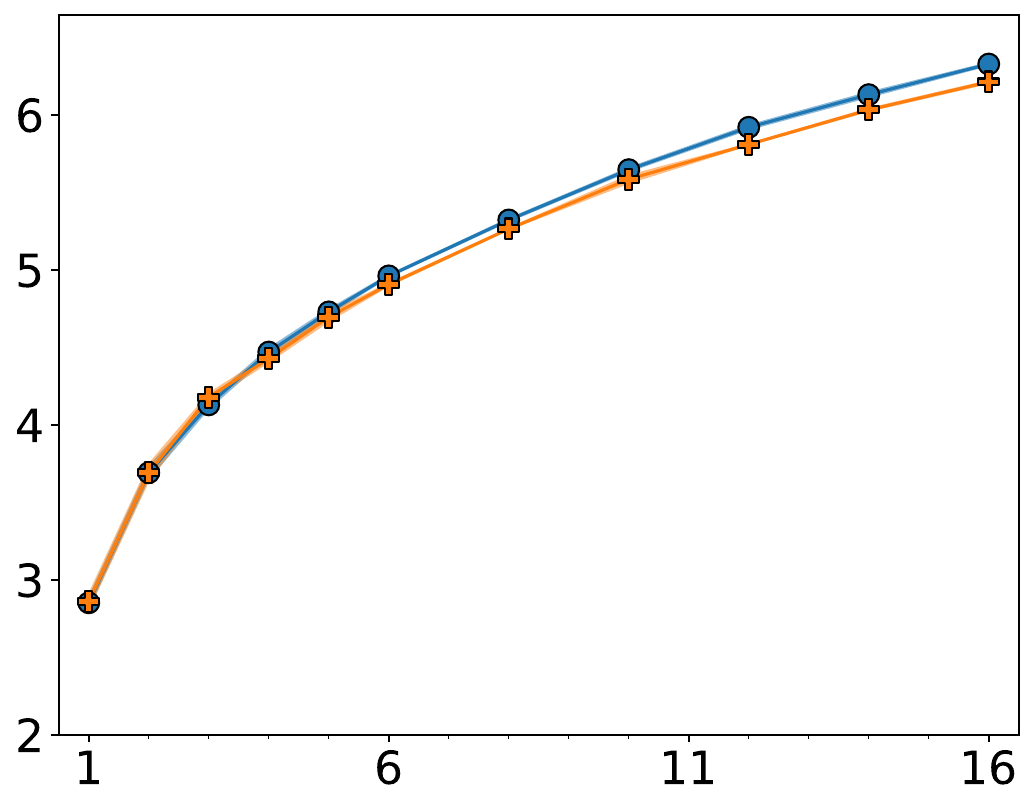}};
  \node at (0.15,-1.6) {\tiny Latent dimension};
  \node[anchor=center,rotate=90] at (-2.1,0) {\tiny Bits};
\end{tikzpicture}
  \caption{Rate}
\end{subfigure}
\begin{subfigure}[b]{0.3\textwidth}%
\centering%
\begin{tikzpicture}
  \useasboundingbox (-2.2,-2) rectangle (2.2,2);
  \node at (0,0) {\includegraphics[width=4cm]{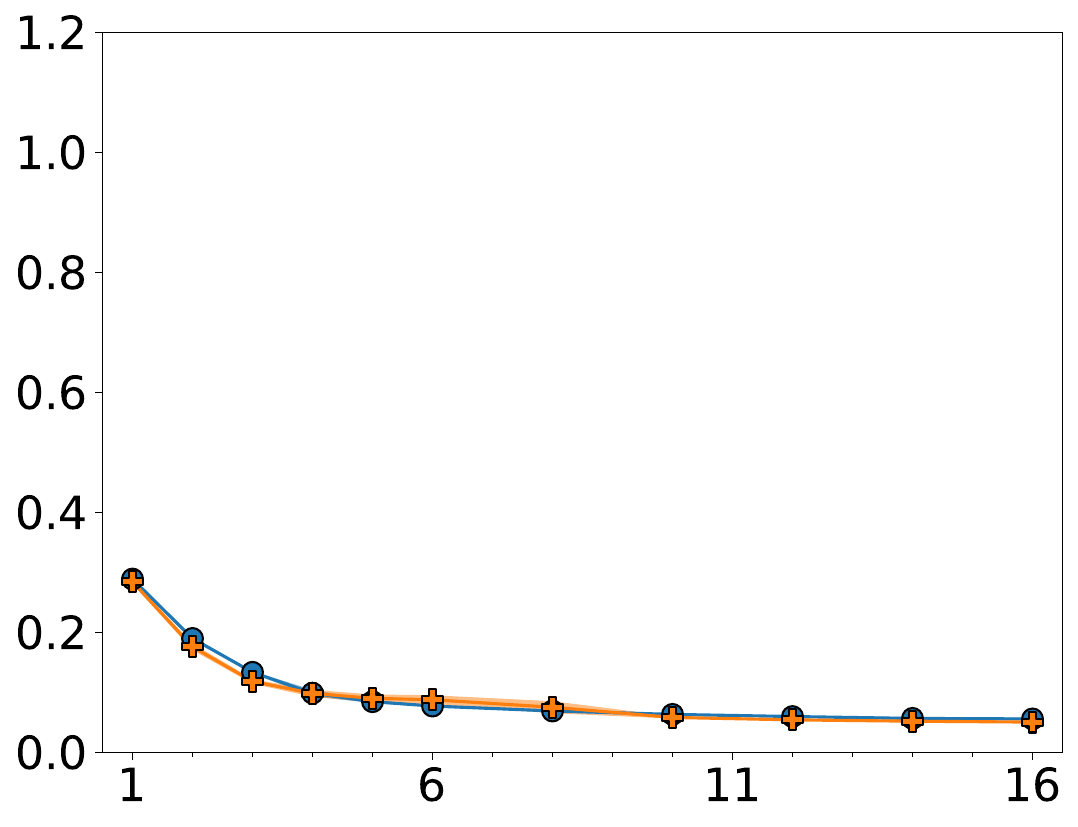}};
  \node at (0.15,-1.6) {\tiny Latent dimension};
  \node[anchor=center,rotate=90] at (-2.1,0) {\tiny Aligned RMSE};
\end{tikzpicture}
  \caption{Train Distortion}
\end{subfigure}
\begin{subfigure}[b]{0.3\textwidth}%
\centering%
\begin{tikzpicture}
  \useasboundingbox (-2.2,-2) rectangle (2.2,2);
  \node at (0,0) {\includegraphics[width=4cm]{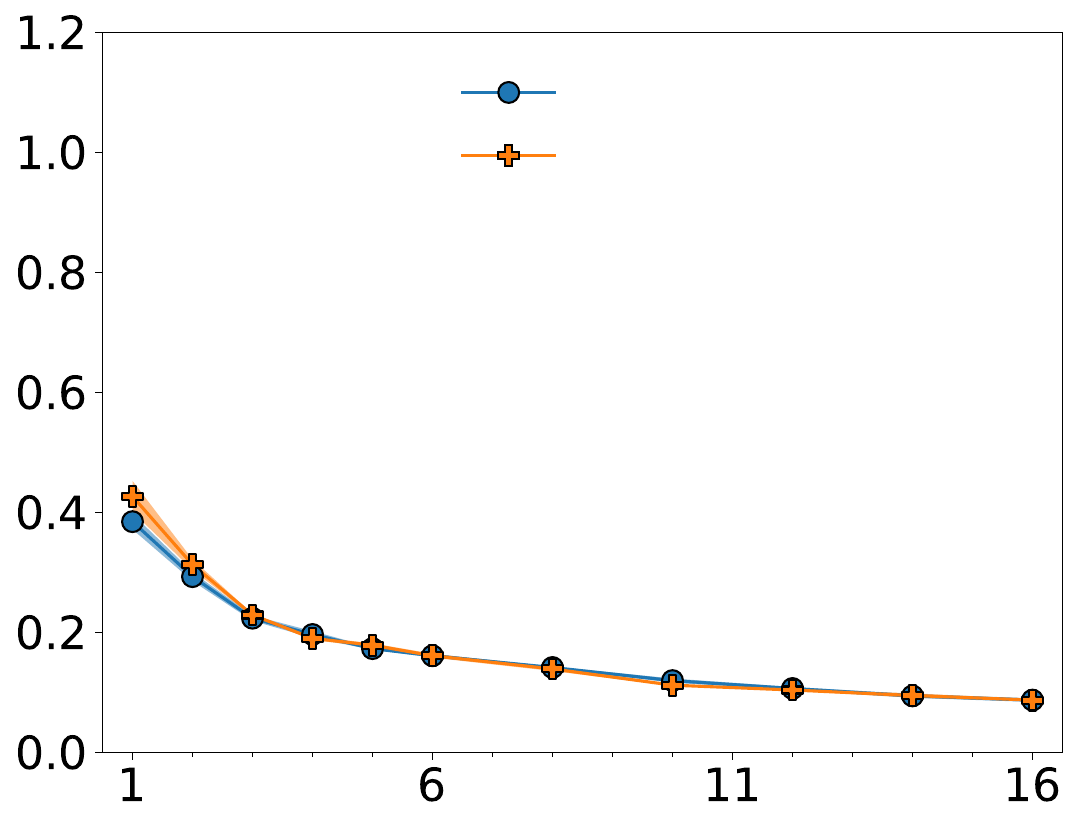}};
  \node at (0.15,-1.6) {\tiny Latent dimension};
  \node[anchor=center,rotate=90] at (-2.1,0) {\tiny Aligned RMSE};
  \def\lx{0.0}
  \def\ly{1.38}
  \node[anchor=north west,font={\fontsize{5pt}{8}\selectfont}] at (\lx,\ly) {TimewarpVAE};
  \node[anchor=north west,font={\fontsize{5pt}{8}\selectfont}] at (\lx,\ly-\lsep) {TimewarpVAE-DTW};
\end{tikzpicture}
  \caption{Test Distortion}
\end{subfigure}
  \caption{Performance comparison between TimewarpVAE and TimewarpVAE-DTW}
  \label{fig:compareDTWResult}
\end{figure}

An alternative formulation of TimewarpVAE, which we call TimewarpVAE-DTW,
is to collect the decoded trajectory as a vector by running the decoder $f(s,z)$ over multiple, 
evenly-sampled canonical timesteps $s$, and then warping those generated timesteps to the training data using
DTW.
This is more similar to Rate Invariant AE, in that the time-warping happens after trajectory generation,
however, we do not require linear interpolation.
Instead, we convert the DTW alignment into a loss in such a way that all canonical trajectory points are used and averaged
(using the same weightings we descriped for our Aligned RMSE).
This avoids the problem encountered in Rate Invariant AE where parts of the canonical trajectory can be completely ignored.
TimewarpVAE-DTW requires running DTW to align each reconstructed trajectory to its associated training trajectory
every time the decoder function is executed during training. 
This is significantly less efficient than our suggested implementation of TimewarpVAE, 
because it requires many executions of dynamic time warping
with no re-use of the DTW results between training steps. 
Our suggested implementation, TimewarpVAE,
explicitly models the time-warping, so is able to re-use (and update) the warping function
between steps, rather than recalculating it from scratch each time.
However, we note in Fig.\ref{fig:compareDTWResult}
that TimewarpVAE-DTW, though less efficient, can give comparable results.
In this implementation we use DTW, but a Soft-DTW \cite{cuturi17a} could be used instead.

\label{sec:timewarpVAEDTWResults}

\subsection{Hyperparameters}
\label{sec:hyperparameters}

The specific training architectures we use is shown in Table~\ref{tab:hyperparameters}.
We use a kernel size of 3 for all convolutions. 
$e$ is the spatial encoding architecture (which is always reshaped to a vector and followed by two separate fully connected layers, one outputting the expected encoding, and one outputting the log of the diagonal of the covariance noise).
$h$ is the temporal encoding architecture,
which is always followed by a fully connected layer outputting a vector of size $K=50$.
$g(s)$ is part of the factorized decoder architecture, which is followed by a fully connected layer outputing a vector of size $m=64$.
$\mathbf T(z)$ is the other part of the factorized decoder architecture, which is followed by a fully connected layer outputting a vector of size $nm$ where $n$ is the number of channels in the training data (2 for handwriting, 7 for fork trajectories) and $m = 64$.
For beta-VAE, instead of the factorized decoder architecture,
we use one fully connected layer with output size $800$,
which we then reshape to size $25\times 32$.
This is followed by one-dimensional convolutions.
Following the approach of \cite{convUpsampling}, 
for the convolutions in the beta-VAE architecture,
instead of doing convolutional transposes
with strides to upsample the data, 
we instead always use a stride length of 1
and upsample the data by duplicating each element before performing each convolution.
Thus, the lengths expand from $25$ in the input to the output size of $200$ after the three convolutions.

We use a learning rate of 0.0001, a batch size of 64, and a rectified linear unit (ReLU) for all spatial and temporal encoder nonlinearities, except for Rate Invariant AE for which follow the literature and we use Tanh. We use an exponential linear unit (ELU) for the decoder nonlinearities for TimewarpVAE, we use ReLU for the decoder nonlinearities for beta-VAE, and we again use Tanh for the Rate Invariant AE.
We choose a variance estimate of $\sigma_R^2 = 0.01$ for our data, but this hyperparameter is not critical, as it is equivalent to scaling $\beta$ and $\lambda$ in our TimewarpVAE objective.
In order to compute the Rate (information bottleneck) of the Rate Invariant AE, we implement it as a VAE instead of an autoencoder, only adding noise to the spatial latent,
not to the timing latent values. We use 199 latent variables for the timing (one fewer than the trajectory length), and vary the number of spatial latent variables.
\begin{table}
  \centering
  \caption{Hyperparameters}
  \label{tab:hyperparameters}
  \small
  \begin{tabular}{lrrrrr}
    \toprule
    Name & $e$ conv channels & $h$ conv channels& $g(s)$ fc& $\mathbf T(z)$ fc  & $f$ fc\\
         &           strides &           strides&            &              &     conv channels\\
    \midrule
    TimewarpVAE & [16,32,64,32] & [16,32,32,64,64,64] & [500,500] & [200]    & --\\
                & [1,2,2,2]        & [1,2,1,2,1,2]       &            &  & \\
    NoTimewarp  & [16,32,64,32] & -- & [500,500] & [200]    & --\\
                & [1,2,2,2]        &    &            &  & \\
    NoNonlinearity & [16,32,64,32] & [16,32,32,64,64,64] & [500,500] & []    & --\\
                   & [1,2,2,2]        & [1,2,1,2,1,2]       &            &     & \\
    beta-VAE    & [16,32,64,32] & -- & -- & --                                     & [800] \\
                & [1,2,2,2]        &    &    &                                        & [20,20,$n$]   \\
    Rate Invariant AE & [32,32,32] & -- & -- & --                                     & [6400] \\
                                  & [1,1,1]     &    &    &                                        & [32,32,$n$]   \\
  \end{tabular}
\end{table}

\subsection{Robot Execution}
Here we give specifics on the execution of the TimewarpVAE trajectory on the Kinova Gen3 robot arm.

\subsubsection{Optimization Formulation}
We formulate an optimization problem to find the fastest feasible trajectory that satisfies the joint speed an torque constraints of our Kinova Gen3 robot arm and follows the training demonstration trajectory within a defined time-warping cost. We use Drake~\cite{drake} to efficiently formulate several of the constraints, including the forward dynamics of the manipulator arm.

We write the time-warping function $\phi$ now as a function from real time $[0,T]$ to canonical time $[0,1]$.
Given the target canonical path $p(s)\ : \ [0,1] \to \mathbb R^3$, 
and our robot joints as a function of the real time $j(t)\ : \ [0,T] \to \mathbb R^7$
and our forward robot kinematics mapping joint angles to end-effector position
$f(j) \ : \ \mathbb R^7 \to \mathbb R^3$.
Our first constraint is that the robot must follow the canonical path according to the time-warping funciton
\begin{equation}
  p(\phi(t)) = f(j(t))
\end{equation}

We also constrain the time-warping function $\phi$ to be monotonically positive from $[0,T]$ to $[0,1]$ by constraining its slope to be positive and for $\phi$ to have the right boundary constraints:
\begin{equation}
  \phi'(t) > 0
\end{equation}
\begin{equation}
  \phi(0) = 0
\end{equation}
\begin{equation}
  \phi(T) = 1
\end{equation}

We define the forward kinematics funciton $f$ by loading the Kinova Gen3 URDF into Drake.
Likewise, we model the forward dynamics using drake, so that, under input joint torques at time $t$given by $\tau(t) \ : \ [0,T] \to \mathbb R^7$ we know
the joint accelerations $j''(t)$ as a funciton of time.
The forward dynamics gives us a differential equation of $j''(t)$ as a funciton of $j', j,$ and $\tau$. Using $FD$ as the forward dynamics function, we have the constraint
\begin{equation}
  j''(t) = FD(j'(t), j(t), \tau(t))
\end{equation}

We additionally add joint 
speed limits ${j'_i}_\text{max}$ for each joint $i$.
\begin{equation}
  -{j'_i}_\text{max} \le j'_i(t) \le {j'_i}_\text{max}
\end{equation}

We also add torque constraints for each joint $i$
\begin{equation}
  -{\tau_i}_\text{max} \le \tau_i(t) \le {\tau_i}_\text{max}
\end{equation}

Finally, we add a time-warping constraint, that our time-warping regularization value must be less than some bound. We note that we adjust the definition of our time-warping cost to account for the new domain of the time-warping of $[0,T]$ instead of $[0,1]$. Our time-warping regularization value now contains additional factors of $T$ and $\frac 1 T$ so that it is not affected by linear scaling of the real time duration $T$.
\begin{equation}
  \frac 1 T \int_0^T \left(T\phi'(t) - 1\right)\log(T\phi'(t))\, dt
\end{equation}
and we constrain this to be below some value $C_\text{warp}$
\begin{equation}
  \frac 1 T \int_0^T \left(T\phi'(t) - 1\right)\log(T\phi'(t))\, dt \le C_\text{warp}
\end{equation}

Our optimization objective is to minimize $T$ (how long it takes the robot to perform the motion) subject to all the constraints above.

\subsubsection{Optimization Using Direct Collocation}
To numerically solve the optimization problem, we use the direct collocation appraoch of~\cite{Hargraves1987} implemented in Drake.
Rather than requiring that all the constraints above be satisfies at all timesteps, we only check constraints at fixed, equally-spaced timesteps.
In particular, we approximate $j(t)$,$j'(t)$,and $\phi(t)$ as cubic splines (with $K$ segments) and $K+1$ equally spaced knots (including the knots at the endpoints at $0$ and $T$), and we constrain the derivative of the cubic spline $j(t)$ to equal the cubic spline $j'(t)$
at the knot and at the collocation points (which are the midpoints in between two knots).
We restrict $\tau(t)$ and $\phi'(t)$ to be piecewise affine with the same equally-spaced knots, and we now just check the following constraints at
knot points and collocation points:
\begin{equation}
  p(\phi(t)) = f(j(t))
\end{equation}
\begin{equation}
  -{j'_i}_\text{max} \le j'_i(t) \le {j'_i}_\text{max}
\end{equation}
\begin{equation}
  -{\tau_i}_\text{max} \le \tau_i(t) \le {\tau_i}_\text{max}
\end{equation}
\begin{equation}
  j''(t) = FD(j'(t), j(t), \tau(t))
\end{equation}

The following constraints are only checked at knot points:
\begin{equation}
  \phi'(t) > 0
\end{equation}

And, of course, the following constraints are still just at the boundary points:
\begin{equation}
  \phi(0) = 0
\end{equation}
\begin{equation}
  \phi(T) = 1
\end{equation}

The warping cost constraint is now approximated by a trapizoidal integration over the $K+1$ knot points $t_k \in \{0,\ldots,T\}$:
\begin{equation}
 \sum_{t_k} w_k \left(T\phi'(t_k) - 1\right)\log(T\phi'(t_k)) \le C_\text{warp}
\end{equation}
where the weighting $w_k$ is $\frac 1 {2K}$ if $t_k$ is $0$ or $T$ (a boundary) and $\frac 1 K$ otherwise.

We then optimize using IPOPT~\cite{Wchter2005}, terminatng after 2000 max iterations.
We use $K=20$ cubic spline segments.
For NoWarping, we set the max time-warping cost $C_\text{warp}$ to zero, and when we allow warping, we set the max time-warping allowed to $0.05$. 
Based on the Kortex Gen3 User Guide, we set the joint speed limits to 1.39 rad/s for the first four joints and 1.22 rad/s for the last three joints.
Likewise, we set the joint torque limits to 32 Nm for the first four joints and 13 Nm for the last three joints.

\subsubsection{Robot Control}
For robot execution, we send low-level joint commands at 500Hz to the Kinova Gen3 arm, using a PID controller to follow the desired joint position and velocities.
The low-level commands directly specify the current desired for each joint, since we found there to be an unusably high delay when commanding via the TORQUE\_HIGH\_VELOCITY low-level API provided by Kinova, leading to controller instability.
The TORQUE low-level API is implemented by Kinova leads to stable control, but contains a cascading controller that includes a velocity control loop, and therefore does not match the expected relationship between commanded torque and joint accellerations (in particular, commanding a robot joint to move with maximal torque accelerates the joint much more slowly than it would without Kinova's cascading controller).
For these reasons, we use the MOTOR\_CURRENT low-level API in a PID loop to command the Kinova Gen3 arm to track the joint trajectory returned from our optimization calculation.

\end{document}